\documentclass[letterpaper]{article} % DO NOT CHANGE THIS
\usepackage{aaai2026}  % DO NOT CHANGE THIS
\usepackage{times}  % DO NOT CHANGE THIS
\usepackage{helvet}  % DO NOT CHANGE THIS
\usepackage{courier}  % DO NOT CHANGE THIS
\usepackage[hyphens]{url}  % DO NOT CHANGE THIS
\usepackage{graphicx} % DO NOT CHANGE THIS
\usepackage{natbib}  % DO NOT CHANGE THIS AND DO NOT ADD ANY OPTIONS TO IT
\usepackage{caption} % DO NOT CHANGE THIS AND DO NOT ADD ANY OPTIONS TO IT
\nocopyright

\usepackage{amsfonts}
\usepackage{amsmath}
\usepackage{amssymb}
\usepackage{amsthm}

\usepackage[noend,ruled,vlined]{algorithm2e}
\usepackage[tight,scriptsize]{subfigure}

\usepackage{pgffor}

\newcommand{\supp}{the Extended Version}
    
\newcommand{\tm}{\mathit{time}}
\newcommand{\wait}{\mathit{wait}}
\newcommand{\Nset}{\mathbb{N}} 
\newcommand{\Val}{\operatorname{Val}}
\newcommand{\prob}{\mathbb{P}}
\newcommand{\AttackVal}{\mathcal{D}}
\newcommand{\ellattacks}{\mathcal{E}}
\newcommand{\adjustmem}{\textsc{AdjustMemory}}

\DeclareMathOperator{\sign}{sign}
\newcommand\mem{\operatorname{mem}}

\title{Memory Assignment for Finite-Memory Strategies\\in Adversarial Patrolling Games}
\author{
Vojt{\v{e}}ch K\r{u}r,
V\'{\i}t Musil,  
Vojt\v{e}ch \v{R}eh\'{a}k}

\affiliations{
Masaryk University, Brno, Czechia\\
vojtech.kur@mail.muni.cz,
\{musil, rehak\}@fi.muni.cz
}

\begin{document}

\maketitle

\begin{abstract}
\emph{Adversarial Patrolling games} form a subclass of Security games where a \emph{Defender} moves between locations, guarding vulnerable targets.
The main algorithmic problem is constructing a strategy for the Defender that \emph{minimizes} the worst damage an \emph{Attacker} can cause.
We focus on the class of \emph{finite-memory} (also known as regular) Defender's strategies that experimentally outperform other competing classes.
A finite-memory strategy can be seen as a positional strategy on a finite set of states.
Each state consists of a pair of a location and a certain integer value--called memory.
Existing algorithms improve the transitional probabilities between the states but require that the available memory size itself is assigned at each location \emph{manually}.
Choosing the right memory assignment is a well-known open and hard problem that hinders the usability of finite-memory strategies.
We solve this issue by developing a general method that iteratively changes the memory assignment.
Our algorithm can be used in connection with \emph{any} black-box strategy optimization tool.
We evaluate our method on various experiments and show its robustness by solving instances of various patrolling models.
\end{abstract}

\begin{links}
    \link{Code}{https://gitlab.fi.muni.cz/formela/regstar}
    % \link{Datasets}{https://aaai.org/example/datasets}
    \link{Extended version}{https://arxiv.org/abs/2505.14137}
\end{links}

\section{Introduction}

This work follows the \emph{patrolling games} line of work studying non--cooperative games where a mobile Defender guards resources against an Attacker.
In \emph{adversarial} patrolling \citep{VAT:adversarial-patrolling,AKK:multi-robot-perimeter-adversarial,AKKS:adversarial-uncertain,BGA:large-patrol-AI,BGA:patrolling-arbitrary-topologies,MunozdeCote2013,SLESSLIN2019}, the Attacker \emph{knows} the Defender's strategy and observes his locations.
The solution concept is based on \emph{Stackelberg equilibrium}.
The Defender commits to a strategy $\sigma$, and the Attacker chooses a counter-strategy $\pi$ that maximizes the expected Attacker's utility against $\sigma$.
Intuitively, the value $\Val(\sigma)$ denotes the damage caused by the best Attacker, and the precise definition of $\Val$ depends on the underlying \emph{patrolling model}.

The environment where the agents play can be described by a Markov decision process.
If the environment is not fully known, the problem becomes one of reinforcement learning, where the agent learns the strategy through interaction with the environment, using either model-free (e.g., Q-learning) or model-based (e.g., using a learned model to simulate outcomes) approaches.
We focus on the cases when the \emph{environment is fully known} both to the Defender and the Attacker.
An example of such an environment is an oriented graph that typically describes the topology of the terrain.
This concept is also used in the domain of planning.

Another aspect is the time frame in which the game is evaluated, which plays a critical role in shaping the agent’s behavior.
We focus on an \emph{infinite-horizon model}, where the Attacker can wait as long as needed to exploit the best opportunity that maximizes the targets' damage.
This is especially suitable for uninterrupted services where the Defender represents many consecutive patrol shifts or a robot that is expected to run for a long time.

For general topologies, deciding whether there is a Defender's strategy with null expected damage is PSPACE-complete\cite{HO:UAV-problem-PSPACE}.
Additionally, deciding whether there is a strategy with at most $\varepsilon$ expected damage is NP-hard~\cite{KKR:patrol-drones} (where $\varepsilon \leq 1/2n$ and $n$ is the number of vertices).
Therefore, no polynomial algorithm can guarantee (sub)optimality in general (unless $\mathrm{P} = \mathrm{NP}$).
Hence, finding suitable heuristics applicable to real-world scenarios is of great importance.

Prior works focus on constructing \emph{positional strategies} where the Defender makes a randomized choice of the next location based only on the current location
\cite{BGA:large-patrol-AI,ACKL:patrol-in-uniform,BDG:spatially-uncertain,CCGKK:fragmented-boundaries-memoryless}.
However, positional strategies fail to provide optimal protection even in a simple setting; see Figure~\ref{fig-positional_not_opt} or \cite{KL:patrol-regular}.

To increase the strategy expressivity, \citet{KL:patrol-regular} introduced a class of \emph{regular strategies}, where the Defender is equipped with a suitable finite-state automaton.
In each location, the distribution over the next locations is determined by the state of the automaton after feeding it with the current history of locations.
The authors proved that regular strategies offer better protection than window-based strategies, which generalize positional strategies by considering the last $k$ locations.
They also present a strategy synthesis algorithm for regular strategies.
However, this approach's main limitation is that the finite-state automaton must be supplied manually to achieve strong strategies.

The idea of regular strategies was followed by \emph{finite-memory} strategies where the finite-state automaton is abstracted into finite integer memory denoted $M$~\cite{KKLR:patrol-gradient}.
The Defender chooses the new location based on the current location and memory value and also selects (possibly randomly) the new memory value.

Finite-memory strategies can also be seen as positional strategies on the set of states consisting of the pairs of the location and a memory integer between $1$ and $M$.
In this view, the state space increases dramatically, limiting the usability and performance of optimization algorithms.
In fact, the size of the memory does not necessarily need to be the same in each location, i.e.,~the states can be set as $C =\{(v, i) \mid v \in V, 1 \leq i \leq \mem(v)\}$, where $V$ is a set of locations and $\mem \colon V \to \Nset$ is certain \emph{memory-assignment} function.

Recent works provide well-performing optimization algorithms for finite-memory strategies and various patrolling models \cite{KKLR:patrol-gradient,KKMR:Regstar-UAI,BKKMNR:Patrolling-changing-UAI,KKMR:expected-intrusion-aamas}.
However, they all assume that a suitable memory assignment is given as an input.
Currently, this is the main bottleneck of these algorithms, and the results depend heavily on the choice of~$\mem$.

All the published experiments either use uniform memory assignment with increasing $M$ (that hits the limits of the algorithms already for $M\lesssim10$) or use carefully handcrafted assignment $\mem$ based on the expert knowledge of the problem at hand.
To the best of our knowledge, \emph{no approach automates memory assignment} for general topologies.
\citet{KKMR:Regstar-UAI}~state that choosing the optimal set of states is an open and hard problem.
We demonstrate the problem of optimal memory assignment in Example~1 below. 

\paragraph{Example 1}

Consider a graph where three target vertices $v_1, v_2, v_3$ are connected to a central vertex $X$, forming a star topology.

Assume that moving along any edge takes 1 time unit, and each target requires 6 time units to complete an attack.
The Defender uses a patrolling strategy that cycles through the targets in the sequence:
\begin{equation*}
    v_1 \to X \to v_2 \to X \to v_3 \to X \to v_1.
\end{equation*}
Since the length of this cycle is $6$, the patrol takes 6 time units to complete.
This way, the Defender reaches each target within the attack duration and can thus intercept any ongoing attack in time.

However, this strategy is not positional since the Defender’s decision at $X$ depends on the history.
We can model this behavior as a finite-memory strategy, where the Defender must maintain three distinct internal states at $X$, denoted $ X_1, X_2, X_3$, resulting in the sequence:
\begin{equation*}
    v_1 \to X_1 \to v_2 \to X_2 \to v_3 \to X_3 \to v_1.
\end{equation*}

Now suppose that the attack duration of $v_1$ is $4$ and the attack durations of $v_2$ and $v_3$ are $8$.
The original strategy is no longer good, since an observing and patient Attacker can wait for the moment when the Attacker moves from $v_1$ to $X$ and initiate an attack on $v_1$.
Since the Defender will return to $v_1$ after $6$ time units, the attack will be completed.

In this case, the optimal strategy returns to $v_1$ after each visit of $v_2$ and $v_3$, resulting in a sequence:
\begin{equation*}
    v_1 \to X \to v_2 \to X \to v_1 \to X \to v_3 \to X \to v_1.
\end{equation*}
Here, the Defender needs two different states for $v_1$ and $4$ for $X$.
Both settings and their respective optimal strategies are visualized in
Figure~\ref{fig-deterministic_star}.

\emph{Discussion.}
This example demonstrates the difficulty of choosing the optimal memory assignment.
It shows that the minimal optimal set of states does not solely depend on the topology but differs based on particular properties of the targets.
Moreover, memory may serve purposes beyond merely remembering the last visited vertex; it can capture more complex patterns in the Defender's behavior.

\begin{figure}[tb]
\centering
\includegraphics[width=\columnwidth]{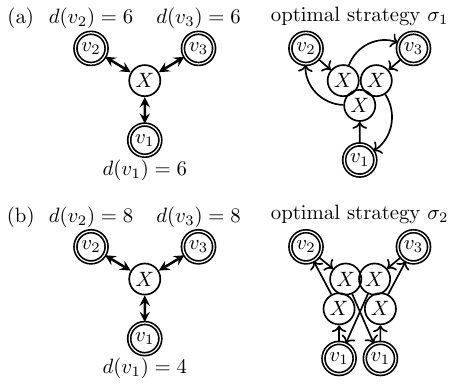}
\caption{The optimal deterministic strategies require different memory allocations.
The goal is to patrol a graph (a) with three targets and an internal location $M$. It takes $d = 6$ time units to complete an attack on each target. The optimal strategy is to cycle around the graph since the cycle's length is $6$. This is represented as $\sigma_1$, which requires $3$ different states at $M$.
In (b), the attack durations change. The strategy $\sigma_2$ achieves perfect protection since $v_1$ is always visited every $4$ steps, and both $v_2$ and $v_3$ are visited once on the cycle of length 8. Strategy $\sigma_2$ needs $4$ states in $M$ and $2$ states in $v_1$.}
\label{fig-deterministic_star}
\end{figure}

\subsection*{Our Contribution}

We propose a smart general heuristic algorithm that dynamically adjusts the memory assignment for finite-memory strategies without any expert knowledge.
Our method works with any differentiable value function and any black-box strategy optimization tool.
In combination, our memory assignment algorithm, together with any strategy synthesis algorithm, yields a general and fully automated method for a strong finite-memory Defender's strategy.

We evaluate our method on several benchmarks and for various patrolling models.
Our automated approach outperforms the manual memory assignment when $\mem$ is chosen to be uniform with increasing $M$ in nearly all cases.
Furthermore, our method is on par with expertly handcrafted memory assignments.

\subsection*{Related Work}

This work contributes to the field of security games, which focuses on the optimal allocation of limited security resources to achieve effective target coverage, see monograph~\cite{Tambe:book}.
Patrolling games are a specialized type of security games where the Defender is mobile; see survey works~\cite{HZHH:multi-robot-patrol-survey,ARSTMCC:multi-patrolling-survey,PR:multi-patrolling-survey,Basilico2022:recent-trends-in_rob_patrol}. 
Most existing patrolling models can be categorized as either \emph{regular} or \emph{adversarial}.

\emph{Regular patrolling} \cite{non-adversarial-patrolling1,non-adversarial-patrolling2,non-adversarial-patrolling3} is akin to surveillance, where the Defender's goal is to quickly discover incidents by minimizing the time between consecutive visits to each target. 
In this model, the Defender typically employs a strategy involving a single path or cycle that visits all targets.

In contrast, \emph{adversarial patrolling} \cite{VAT:adversarial-patrolling,AKK:multi-robot-perimeter-adversarial,AKKS:adversarial-uncertain,BGA:large-patrol-AI,BGA:patrolling-arbitrary-topologies,MunozdeCote2013,SLESSLIN2019,Basilico2022:recent-trends-in_rob_patrol} focuses on protecting targets from an Attacker who seeks to exploit the best opportunities to maximize damage. 
This model is generally framed within the context of Stackelberg equilibrium \cite{YKKCT:Stackelberg-Nash-security,SFAKT:Stackelberg-Security-Games}.
The Defender's strategies are often \emph{randomized} to prevent the Attacker from predicting future moves, and the Defender aims to maximize the probability of detecting an attack.
The adversarial approach is particularly relevant in scenarios where a certain level of protection must be maintained, even if incidents occur at the most inconvenient times.

For patrolling scenarios of bounded duration, such as an eight-hour shift of a human ranger, \emph{finite-horizon} models are sufficient.
In contrast, \emph{infinite-horizon} models are used when the duration is potentially unbounded or the bound is large, like 24/7 surveillance.

The model can also be distinguished by prior knowledge about the environment, which can be either fully known to all players or needs to be discovered during patrolling.

The difference between models
has a substantial impact on strategy synthesis techniques.
For finite horizon models with known environments, mathematical programming is the primary technique \cite{BGA:large-patrol-AI,BGA:patrolling-arbitrary-topologies}.
Reinforcement learning has been largely successful for patrolling scenarios with a finite horizon and unknown environment, such as green security games \cite{WSYWSJF:Patrolling-learning,BAVT:learn-preventive-healthcare,Xu:Green-security,KMZGA:moving_targets}. 
Gradient descent methods for finite-memory strategies are used extensively for infinite horizon models with known environments \cite{KKLR:patrol-gradient,KKMR:Regstar-UAI,BKKMNR:Patrolling-changing-UAI,KKMR:expected-intrusion-aamas}.

Practical applications of security games include the deployment of police checkpoints at Los Angeles International Airport \cite{PJMOPTWPK:Deployed-ARMOR}, the scheduling of federal air marshals on domestic airline flights across the U.S. \cite{TRKOT:IRIS}, and the strategic arrangement of city guards in the Los Angeles Metro \cite{DJYZTKS:patrolling-uncertainty-JAIR}, positioning of U.S. Coast Guard patrols to secure critical locations \cite{ASYTBDMM:Protect-AIMagazine}, as well as the initiatives for wildlife protection in Uganda \cite{FKDYT:PAWS}.

\section{Background}

First, we recall the standard notions of a patrolling graph, the Defender's and Attacker's strategy, and their value, and we fix the notation.
Then, we formulate the requirements of our algorithm for the underlying patrolling model
and show how existing patrolling models fit our framework.
We assume familiarity with basic notions of calculus, linear algebra, probability theory, and finite-state discrete-time Markov chains. 

\subsection{Patrolling Graph}
A \emph{patrolling graph} is a tuple $G = (V,T,E, \tm)$, where $V$ is a non-empty set of \emph{locations} (admissible Defender's positions), $T \subseteq V$ is a non-empty set of \emph{targets}, $E \subseteq V \times V$ is a set of \emph{edges} (admissible Defender's moves) and $\tm\colon E\to\Nset$ specifies the traversal time of an edge.

\subsection{Defender's Strategy}
In general, the Defender can choose the next location randomly based on the whole history of previously visited locations.
However, general strategies may not be finitely representable.
We focus on finite-memory Defender's strategies, which were shown to achieve the same limit protection as general strategies in two different patrolling models; see \cite{KKMR:Regstar-UAI} and \cite{KKMR:expected-intrusion-aamas}.

In finite-memory strategies, the Defender is equipped with an integer variable $M$ called \emph{memory}.
A \emph{state} of the Defender is a pair $(v, i) \in V \times \Nset$.
A finite-memory strategy of the Defender can be seen as positional on a fixed finite set of states, i.e., it forms a discrete-time Markov chain on $V \times \Nset$.

To prevent the state space from blowing up, different amounts of memory can be allocated to each location.
Hence, we define the \emph{memory assignment} as a function $\mem \colon V \to \Nset$ and a set of Defender's states
\begin{eqnarray}
    C =\{(v, i) \mid v \in V, 1 \leq i \leq \mem(v)\}.
\end{eqnarray}
A finite-memory strategy of the Defender on $V$ is a positional strategy on $C$, i.e., a function $\sigma \colon C \times C \to [0, 1]$ satisfying $\sum_{y \in C}\sigma(x,y) = 1$, for each $x \in C$, and $(u, v)\in E$, whenever $\sigma((u, i), (v, j)) > 0$.
Intuitively, if the Defender is currently in a location $u$ with a memory value $i$, the next state $(v, j)$ is chosen with probability $\sigma((u, i), (v, j))$, which means that the Defender changes the memory value to $j$ and starts traveling from $u$ to $v$.
The second condition ensures that the strategy respects the edges of the graph.

For every finite sequence of states $h = (c_1, \ldots, c_n)$, we use $\prob^{\sigma, c}\lbrack h \rbrack$ to denote the probability of executing $h$ when the Defender starts patrolling in the state $c \in C$ and follows the strategy $\sigma$.
That is, $\prob^{\sigma, c}\lbrack h \rbrack = 0$ if $c_1 \neq c$, and $\prob^{\sigma, c}\lbrack h \rbrack = \prod_{i = 1}^{n - 1}\sigma(c_i, c_{i + 1})$ otherwise.

\paragraph{Strategy Types}\label{sec-strategy_types}

If $\mem \equiv 1$, then we say that $\sigma$ is positional on $V$ or \emph{memoryless}.
If for every state $x$ there is a state $y$ such that $\sigma(x, y) = 1$, then we say that $\sigma$ is \emph{deterministic}.
Note that $\sigma$ can be both memoryless and deterministic.

\subsection{Attacker's Strategy}

In the patrolling graph, the time is spent traversing the edges. % rather than on the locations.
In adversarial models, the Attacker is assumed to perfectly observe the Defender's moves and can determine the next edge taken by the Defender immediately after its departure.
For the Attacker, this is the best moment to attack because delaying the attack could only lower the attack's gain.
Furthermore, the Attacker is allowed to attack only once in a single run.

An \emph{observation} is a sequence of states $o = (c_1, \cdots, c_n, c_{n + 1})$.
Intuitively, $c_1$ is the initial state of the Defender, $c_n$ is the current state of the Defender, and $c_{n+1}$ is the state chosen as the next one according to the Defender's strategy.
The set of all observations is denoted by $\Omega$.
Formally, an \emph{Attacker's} strategy is a function $\pi \colon \Omega \to T \cup \{\wait\}$.
Since the Attacker is allowed to attack only once, it is required that if $\pi(o) \in T$, then $\pi(o') = \wait$ for every prefix $o'$ of $o$.

\subsection{Protection Value and Target Types}

Let $\sigma$ be a finite-memory Defender's strategy and $\pi$ an Attacker's strategy.
Let us fix an initial state $c$ where the Defender starts patrolling.
The expected damage the Attacker causes by attacking $\tau$ after observing $o$ is denoted as $\AttackVal(o, \tau \mid \sigma)$.
We will now present some examples of the definition of the value $\AttackVal(o, \tau \mid \sigma)$. 

\paragraph{Hard-constrained targets}
One way of defining the damage on targets is to model a scenario where the Attacker needs to perform some action (e.g.~picking a lock) to successfully complete an attack.
If the target is visited in time, the Defender `catches' the Attacker, and the damage is zero. Otherwise, the Attacker `steals' the cost of the target. 
Here, $d(\tau)$ denotes the time it takes to complete an attack on $\tau$ and $\alpha(\tau)$ its cost.
The value $\AttackVal(o, \tau \mid \sigma)$ is then defined as the probability that the Defender does not visit $\tau$ in the next $d(\tau)$ time units multiplied by $\alpha(\tau)$.
We call this type of target \emph{hard-constrained}.
Within the context of finite-memory strategies, 
this model was first introduced by \cite{KKLR:patrol-gradient} and later used in
\cite{KKMR:Regstar-UAI,BKKMNR:Patrolling-changing-UAI}.

\paragraph{Blind targets}
The hard-constrained model can be extended by introducing a level of uncertainty where the ongoing attack is discovered only with probability $\beta(\tau)$ upon each visit of the target.
We call these targets \emph{blinded} and appear in \cite{KKMR:Regstar-UAI,BKKMNR:Patrolling-changing-UAI}.

\paragraph{Linear targets}
Another target type models a situation where the actual damage depends on the time elapsed since initiating the attack till the Defender's visit, e.g., a fire or punching a hole in a fuel tank.
Here, the target $\tau$ is assigned a value $v(\tau)$, which denotes the damage caused for every time unit while the target is under attack.
Hence, the expected damage $\AttackVal(o, \tau \mid \sigma)$ equals the expected time it takes to visit $\tau$ multiplied by $v(\tau)$.
If the probability of visiting $\tau$ after $o$ is zero, then $\AttackVal(o, \tau \mid \sigma) = \infty$.
We call this type of targets \emph{linear}.
It was introduced in~\cite{KKMR:expected-intrusion-aamas}.

\subsection{Game Value}

The \emph{expected damage} caused by the Attacker is defined as
\begin{eqnarray}
    \Val(\sigma, c, \pi) = \sum_{\substack{o \in \Omega, \pi(o) \in T}}\prob^{\sigma, c}\lbrack o \rbrack \cdot \AttackVal(o, \pi(o) \mid \sigma).
\end{eqnarray}
The Defender/Attacker aims to minimize/maximize the damage, and hence, we define
\begin{eqnarray}\label{eq-val_best_response}
    \Val(\sigma)
        = \min_{c \in C} \Val(\sigma, c) = \min_{c \in C}\,
          \sup_{\pi} \Val(\sigma, c, \pi)\ .
\end{eqnarray}

By $\Val(G)$, we denote the limit value of the best Defender's strategy or formally $\Val(G) = \inf_{\sigma} \Val(\sigma)$.

\subsection{Requirements on the Patrolling Model}

For our algorithm, we require two properties of the underlying patrolling model.

Let $\sigma$ be a Defender's strategy and let $BSCC(\sigma)$ denote all subsets of states corresponding to bottom strongly connected components of the Markov chain determined by $\sigma$.
For every $B \in BSCC(\sigma)$, let us denote by $\Val(\sigma \mid B)$ the value
\begin{eqnarray*}
    \max_{\substack{\tau \in T , e \in B \times B , \sigma(e) > 0}} \AttackVal(e, \tau \mid \sigma). 
\end{eqnarray*}

Our first requirement is that $\Val(\sigma)$ can be expressed as
\begin{eqnarray}\label{eq-val_max_attacks}
    \Val(\sigma) = \min_{B \in BSCC(\sigma)} \Val(\sigma \mid B) .
\end{eqnarray}
Note that if $\sigma$ is irreducible, then $\Val(\sigma) = \Val(\sigma \mid C)$.

Secondly, given a state space $C$, we require that for every $e \in C \times C$ and $\tau \in T$ the value $\AttackVal(e, \tau \mid \sigma)$ can be effectively calculated, and that $\AttackVal(e, \tau \mid \sigma)$ is a differentiable function with respect to $\sigma$ whose gradient can also be evaluated.

The proofs that hard-constrained, blinded, and linear targets are eligible for our general framework are provided in the works where they were presented.
In general, it can be seen that a sufficient condition for (\ref{eq-val_max_attacks}) is that the damage to the targets depends only on the current state and the future.
That is, for every observation $o = (c_1, \ldots, c_n, c_{n+1})$ and target $\tau$, it holds
\begin{eqnarray}\label{eq-attack_no_history}
    \AttackVal(o, \tau \mid \sigma) = \AttackVal\bigl((c_n, c_{n+1}), \tau \mid \sigma\bigr)\ .
\end{eqnarray}
See \supp{} for a proof.

\begin{figure}[tb]
    \centering
    \includegraphics[width=\columnwidth]{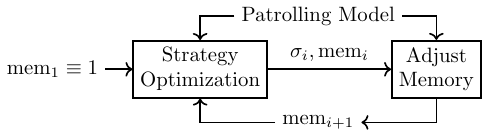}
    \caption{Diagram of our method.
    We iteratively run a strategy optimization black-box followed by our \protect\adjustmem{} procedure that updates the memory assignment.}
    \label{fig:diagram}
\end{figure}

\section{The Method}

Here, we describe our method for solving the problem of choosing a suitable memory assignment.
Our method works with any black-box optimization tool that improves the transitional probabilities and with any patrolling model that fits our general framework.
Recall that the algorithms introduced in \cite{KKMR:Regstar-UAI,BKKMNR:Patrolling-changing-UAI,KKMR:expected-intrusion-aamas,KKLR:patrol-gradient} are eligible instances of such black-boxes.

The algorithm is initialized with the memory assignment $\mem_1$ that assigns $1$ to each location.
That is, we start with a memoryless strategy.
Then, we run the optimization tool and collect the resulting strategy $\sigma_1$.
From this, we run the core procedure \adjustmem{} that outputs a new assignment $\mem_2$.
The memory assignment $\mem_2$ is then used for initialization for the tool, and we collect a new strategy $\sigma_3$. 
This process continues for as long as the strategy value improves, see Figure~\ref{fig:diagram}.

\subsection{Attacks Profiles}
Now, we describe the core procedure \adjustmem{} of our algorithm that decides how much memory should be added to which location.
Intuitively, the procedure estimates the number of different `behaviors' of each state that, on their own, could improve the current value. 

\begin{figure}[tb]\centering
\includegraphics[width=\columnwidth]{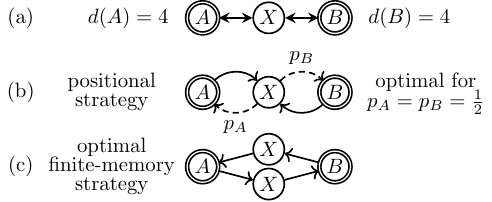}
\caption{Patrolling on a graph (a) with three locations. End locations represent targets with costs of $1$ and attack lengths equal to $4$. Positional strategies (b) are parametrized by a single variable $p$. Due to symmetry, the optimality is reached for $p=0.5$.
The optimal strategy (b) is deterministic and needs two states in $X$.}
%\Description{positional are not optimal.}
\label{fig-positional_not_opt}
\end{figure}

Let us illustrate the idea with a simple example.
Consider a patrolling graph depicted in Figure~\ref{fig-positional_not_opt} with two targets $A$ and $B$ that are hard-constrained with $d = 4$ and $\alpha = 1$.
We start with $\mem = 1$.
The state space is $C=\{ (A, 1), (X, 1), (B, 1) \}$.
For brevity, let us write $v$ for $(v, 1)$ since the memory value is constant.
In this example, from the end locations $A$ and $B$, the Defender always returns to $X$,
and hence the strategy depends only on complementary probabilities $\sigma(X,A)$ and $\sigma(X,B)$,
which we denote by $p_A$ and $p_B$, for short.

The best moment to attack $A$ is when the Defender starts traversing from $X$ to $B$.
Then the probability of not visiting $A$ in $d(A) = 4$ time units is exactly $p_B$ and hence $\AttackVal((X, B),\, A \mid \sigma) = p_B$.
Analogically, the best moment to attack $B$ is when the Defender starts traversing from $X$ to $A$ and  $\AttackVal((X, A),\, B \mid \sigma) = p_A$.
The value $\Val(\sigma)$ is then the maximum of both the attack values, i.e.~$\Val(\sigma)=\max \{p_A, p_B\}$ which is minimized when $\sigma(X,A)=\sigma(X,B)=0.5$ with $\Val(\sigma) = 0.5$.

We have reached the optimal value of positional strategies, and the question is whether the Defender can do better.
The answer is positive since a finite deterministic strategy that moves from one end to the other and back represents a cycle of length $4$.
Hence, every attack is discovered in time, achieving the optimal value $0$.
This strategy can be represented by a finite-memory strategy on memory assignment $\mem_{\mathrm{det}}$ with $\mem_{\mathrm{det}}(X) = 2$ and $\mem_{\mathrm{det}}(A) = \mem_{\mathrm{det}}(B) = 1$.

In practice, a reasonable optimization tool should output a strategy close to $\sigma$.
Now, the key question is how to infer $\mem_{\mathrm{det}}$ from $\sigma$.
The idea is to consider gradients of all the maximal attacks w.r.t.~the parameters of strategy $\sigma$.
Any strategy $\sigma$ can be modeled as Softmax of unconstrained parameters.
In our example, we need just two real parameters, say $(x_A, x_B)$, so that
\begin{eqnarray*}
    \operatorname{Softmax}(x_A, x_B) = (p_A, p_B).
\end{eqnarray*}
Using the chain rule, the partial derivatives of the maximal attacks w.r.t.\ the parameters $x_A$, $x_B$ are
\begin{eqnarray*}
    \tfrac{\partial\AttackVal\bigl((X, B), A \mid \sigma\bigr)}{\partial x_A}  
        = - p_A p_B
    \quad\text{and}\quad
    \tfrac{\partial\AttackVal\bigl((X, B), A \mid \sigma\bigr)}{\partial x_B}
        = p_A p_B
\end{eqnarray*}
and
\begin{eqnarray*}
    \tfrac{\partial\AttackVal\bigl((X, A), B \mid \sigma\bigr)}{\partial x_A}  
        = p_A p_B
    \quad\text{and}\quad
    \tfrac{\partial\AttackVal\bigl((X, A), B \mid \sigma\bigr)}{\partial x_B}
        = - p_A p_B.
\end{eqnarray*}
We can see that the problem lies in the fact that for the attack on $A$, the gradient of the parameters is (up to a scale) $(-1, 1)$, while for the attack on $B$ it is $(1, -1)$, which is in the exact opposite direction.
Hence, the value can no longer improve since the competing gradients `cancel each other out'.
This is exactly the place where we need to add a more memory value for $X$, one memory value to prioritize $B$ (if the previous location was $A$) to cover the attack on $B$, and the second memory value to go to $A$ (if the previous location was $B$) in order to cover the attack on $A$.

Formally, we fix a set of \emph{eligible attacks} $\ellattacks \subseteq (C \times C) \times T$, corresponding to the attacks with close to maximal value.
For an attack $(e,\tau) \in \ellattacks$ and a state $c \in C$, we consider the gradient of the attack value with respect to the outgoing probabilities of $c$
\begin{eqnarray}
    \nabla_\sigma \AttackVal(e, \tau \mid \sigma)
        = \left(\frac{\partial\AttackVal(e, \tau \mid \sigma)}{\partial \sigma(c, c_1)} ,\ldots, \frac{\partial\AttackVal(e, \tau \mid \sigma)}{\partial \sigma(c, c_n)}  \right),
\end{eqnarray}
where $c_1, \ldots, c_n$ are all the states with $\sigma(c, c_i) > 0$.
Assuming that the strategy probabilities are parametrized by the Softmax of $x(c) =(x_1,\ldots,x_n)$,
the gradient of the attack value with respect to the strategy parameters is given by
\begin{equation*}
    \nabla_{x(c)} \AttackVal(e, \tau \mid \sigma)
        = \bigl(\tfrac{\partial\AttackVal}{\partial x_1} ,\ldots, \tfrac{\partial\AttackVal}{\partial x_n} \bigr)
        = J^T\!\cdot\! \nabla_\sigma \AttackVal(e, \tau \mid \sigma),
\end{equation*}
where $J$ is the Jacobian of the Softmax.
The \emph{attack profile} of $(e, \tau)$ and $c$ is then the \emph{sign} of this vector $\nabla_{x(c)} \AttackVal(e, \tau \mid \sigma)$.
We define $\mathrm{profiles}$ to be a function that assigns each state $c$ the set of all (different) attack profiles belonging to $c$ from $\ellattacks$.
Formally, 
\begin{eqnarray*}
    \mathrm{profiles}(c) = \{ \mathrm{sign}( \nabla_{x(c)} \AttackVal(e, \tau \mid \sigma)) \mid (e, \tau) \in \ellattacks\} .
\end{eqnarray*}
The new memory assignment $\mem_2$ is then defined as
\begin{eqnarray}
    \mem_2(v) = \sum_{1 \leq i \leq \mem_1(v)} |\mathrm{profiles}(v, i)|.
\end{eqnarray}

The whole process is also described in the Algorithm~\ref{alg:mem}.

Theoretically, a locally optimal strategy implies the existence of multiple maximal attacks with identical values (conflicting gradients). However, since numerical optimization tools may converge to solutions where these values differ, the threshold $\varepsilon$ ensures we capture all relevant attacks that should be driving the gradient descent.
We have experimentally found that $\varepsilon=0.25$ reliably incorporates all such maximal attacks across different patrolling graphs while significantly reducing computation time.

\subsection{Bounded Number of States}

We also introduce a modification of our method for the cases when the total number of states is bounded above by $L$.
Bounding the maximal number of states of the strategy may be both desired by the user (memory of a robot) and the whole optimization process (algorithms run very slowly for strategies with a high number of states).

Assume we have an upper bound $L \geq |V|$ on the number of states for each $\sigma_i$.
Let the current memory assignment be $\mem_1$ and $\sigma$.
We run \adjustmem{} and collect the resulting $\mem_2$ and attack profiles.
If $\sum_{v}\mem_2(v) \leq L$, we simply output $\mem_2$.
Otherwise, for each attack profile, we sum the total value of the attacks with the given profile.
That is, for each state $c$ and an attack profile $\gamma \in \mathrm{profile}(c)$, we define the \emph{value} of $\gamma$ as 
\begin{eqnarray*}
    \mathrm{val}(\gamma, c) =\sum_{\substack{(e, \tau) \in \ellattacks \\ \sign(\nabla_{x(c)} \AttackVal(e, \tau \mid \sigma)) = \gamma}}\AttackVal(e, \tau \mid \sigma).
\end{eqnarray*}

Then, the idea is to take $L$ profiles with the highest value, ensuring every state has at least one profile.
The memory is then calculated in the same way as in the unbounded version.

Formally, let $\mathrm{profiles'}(c)$ be the set of profiles of $c$, excluding one with maximal value (it does not matter which one).
This is valid since every state must have a profile with an attack of maximal value.
Then we flatten all the attack profiles into a single set $\mathcal{P} = \bigcup_{c}\{(\gamma, c) \mid \gamma \in \mathrm{profiles'}(c)\}$.
After that, we take the $L - |C|$ highest items from $\mathcal{P}$ where the order is given first by $\mathrm{val}$ and secondly by the order on the states (if more profiles of the same state have the same value, the order does not matter).
Let us denote by $\#(c)$ the number of occurrences of a state $c$ among the $L - |C|$ maximal items of $\mathcal{P}$.
Finally, the new memory assignment $\mem_2$ is
\begin{eqnarray*}
    \mem_2(v) = \sum_{1 \leq i \leq \mem_1(v)} 1 + \#(v, i) .
\end{eqnarray*}

\subsection{Complexity Analysis}

Here, we describe the complexity analysis of our method, its bounded variation and strategy expansion.

\paragraph{Algorithm 1}
The time complexity of Algorithm 1 is in
\begin{equation*}
    \mathcal{O}(f + |\ellattacks| g + {|\ellattacks|}^2 {|C|}^2)    
\end{equation*}

where $f$ is such that the values of all attacks are calculated in $\mathcal{O}(f)$ and $g$ is such that for each attack, the gradients with respect to the parameters are evaluated in $\mathcal{O}(g)$.
Next, $|E|$ denotes the number of state edges such that its traversal probability is positive, and finally, $|\ellattacks|$ denotes the number of attacks taken into account.

\paragraph{Bounded case} Let $M = \sum_{v}\mem_2(v)$. If $M \leq L$, then the algorithm is the same as the normal case.
Otherwise, we claim that in runs in
\begin{equation*}
    \mathcal{O}\left(|C||\ellattacks| + (M - |C|)\log(L - |C|)\right) .
\end{equation*}

See \supp{} for proofs.
\emph{In practice}, however, the evaluation for hard-constrained and blinded was faster than for linear. This is primarily because the first algorithm's core procedure is implemented in C++ with a dynamic program and uses a sparse representation of the strategy, unlike the second, which is in PyTorch and uses full matrices. In both cases, the actual bottleneck was in calculating the gradients.

\subsection{Warm-starting}

Some optimization tools enable us to warm-start the process instead of always starting from random parameters.
In \supp{}, we present a way to transform a strategy $\sigma_i$ on $\mem_i$ to $\sigma_{i+1}$ on $\mem_{i+1}$.
However, we conclude that this does not provide a significant advantage, and we omit this extension from the experiments presented here.

\begin{algorithm}[tb]
\caption{Core procedure of our method.}
\label{alg:mem}
\SetAlgoLined
\SetKwFunction{FMain}{AdjustMemory}
\SetKwProg{Fn}{Function}{:}{}
\Fn{\FMain{$\sigma$, $\mem$, $\varepsilon$}}{
    Define $x(c)(c_i)$ to be $\log(\sigma(c, c_i))$\\
    \ForEach{$\mathrm{c} \in \mathrm{C}$}{
        profiles(c) $\gets$ $\emptyset$
    }
    % profiles[c] $\gets$ $\emptyset$ \;
    \ForEach{$e \in C\times C, \tau \in T $}{
        \If{$\AttackVal(e, \tau \mid \sigma) < (1 - \varepsilon)\Val(\sigma)$}{
            \textbf{continue}
        }
        \ForEach{$\mathrm{c} \in \mathrm{C}$}{
            profile $\gets$ $\sign(\nabla_{x(c)} \AttackVal(e, \tau \mid \sigma))$\\
            Add $\mathrm{profile}$ to $\mathrm{profiles}(c)$
        }
    }
    \ForEach{$v \in V$}{
        $\mem'(v)$ $\gets$ 0\\
        \For{$i \gets 1$ \KwTo $\mem(v)$}{
            $\mem'(v)$ $\gets$ $\mem'(v) + |\mathrm{profiles}(v, i)|$
        }
    }
    \Return $\mem'$
}
\end{algorithm}

\section{Experiments}\label{chapter_experiments}

We evaluate our memory assignment method in a series of experiments with the best-performing algorithm for hard-constrained and blind targets of \cite{BKKMNR:Patrolling-changing-UAI} and the algorithm for linear targets of \cite{KKMR:expected-intrusion-aamas}.
We created a unified Python project containing all compatible state-of-the-art algorithms with existing relevant benchmarks.
The repository will be publicly released (preliminarily, see \supp{}).

We also test the manual assignment \texttt{deg} that sets $\mem(v)$ to the out-degree of $v$.
This is precisely the memory needed when an Eulerian cycle forms the optimal strategy.
This memory was expertly handcrafted by the authors \cite{KKMR:expected-intrusion-aamas} for their experiments.

We compare uniform memory assignments (denoted by a single number) with the degree assignment (\texttt{deg}), and our automatic method (\texttt{auto}).
We ran experiments on three patrolling models/target types (hard-constrained, blind, linear) on over 50 different patrolling graphs (Offices, Building, Airports, Terrains, Stars).

For each setting, we executed 200 runs with a timeout of 180 seconds to get reliable statistical information.
The number of states is bounded by 300.
Recall that the whole procedure is stochastic as the optimization starts from a random init and may also contain non-deterministic updates.

Here, we present only a fragment of our experiments to highlight our key observations and limitations.
The broader summary is in Section~\ref{sec:discussion}.
The full report, detailed setup, and reproducibility steps can be found in \supp{}.

\begin{figure}[tb]
    \centering
    \includegraphics[width=1\columnwidth]{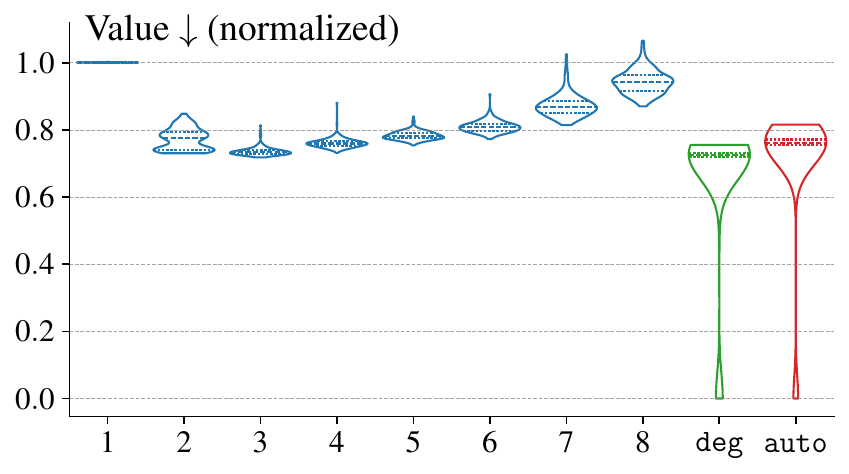}
    \caption{Comparison of memory assignments on the \textbf{Offices} benchmark (2 floors).
    %The values are normalized by the value of the best memoryless strategy, optimum is zero.
    Statistics are over 200 restarts.
    Uniform $\mem = 4$ is sufficient for the optimal strategy, but never found for any of the uniform within the timeout.
    Both Expert (\texttt{deg}) and Automatic (\texttt{auto}) memory assignments lead to the optimum.}
    \label{fig:office_floor_2}
\end{figure}

\paragraph{Offices}
\begin{figure}[tb]
    \centering
    \includegraphics[width=1\columnwidth]{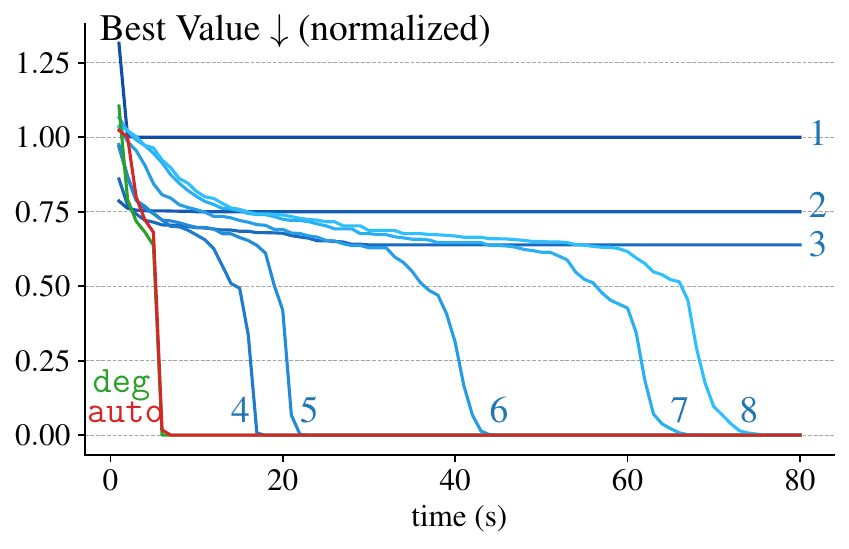}
    \caption{Values over time of the best performing strategies (out of 200 restarts) on one-floor \textbf{Offices}.
    The values are normalized by the value of the best memoryless strategy, optimum is zero.
    Uniform memory below 4 cannot achieve optimum; higher memory gets zero but significantly later as the memory increases.
    Both Expert (\texttt{deg}) and Automatic (\texttt{auto}) outperform the uniform assignment.}
    \label{fig:offices_time_plot}
\end{figure}
This benchmark comes from \cite{KKMR:Regstar-UAI}.
The goal is to patrol office buildings with 1, 2, or 3 floors.
The offices are designed so that an Eulerian cycle is the optimal strategy.
The maximum degree is 4; hence, the uniform memory must be at least 4 to achieve optimum.
%We ran the experiments for offices on $1$, $2$, and $3$ floors.
Figure~\ref{fig:offices_time_plot} shows the optimization dynamics on 1~floor
and
Figure~\ref{fig:office_floor_2} reports the values of strategies for offices on 2~floors.

\paragraph{Building}
\cite{KKMR:Regstar-UAI} also used the same topology as Offices to evaluate the synthesis of blinded targets.
We recreate this benchmark, and the results can be found in \supp{}.

\begin{figure}[bt]
    \centering
    \includegraphics[width=\columnwidth]{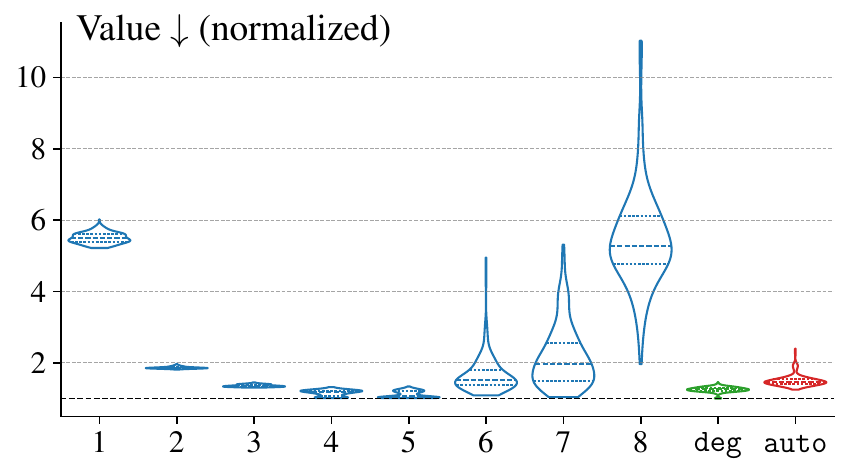}
    \caption{Comparison of memory assignments on the \textbf{Airport} with 19 halls.
    The values are normalized by the value of the baseline (Eulerian cycle requiring degree memory).
    Uniform memory (for $\mem=4$) is sufficient to meet the Expert (\texttt{deg}) assignment.
    This is one of the rare examples where the Automatic (\texttt{auto}) assignment lags behind.}
    \label{fig-airport}
\end{figure}

\paragraph{Airports}
This benchmark originated from~\cite{KKMR:expected-intrusion-aamas}.
The goal is to patrol airport gates.
An airport with $n$ halls has $3n + 1$ nodes, from which $2n$ are linear targets.
In Figure~\ref{fig-airport}, we report the distribution of strategy values for $n=19$.
We also ran the benchmark with random target values sampled uniformly and independently between 1 and~10.

\begin{figure}[tb]
    \centering
    \includegraphics[width=\columnwidth]{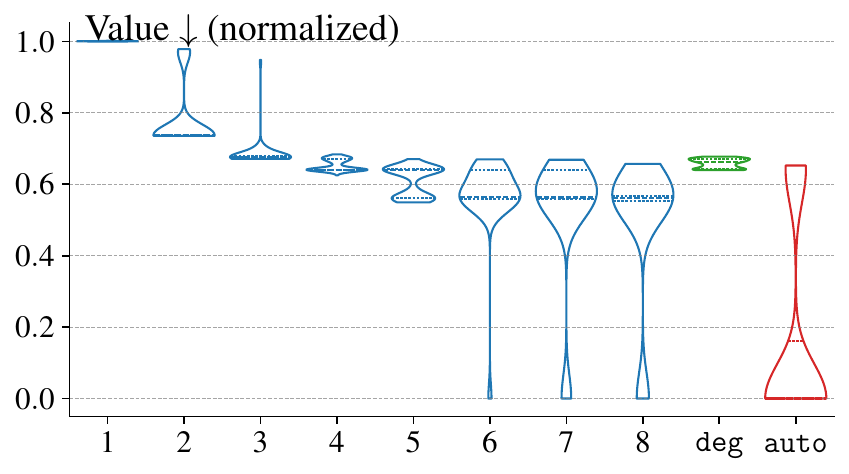}
    \caption{Comparison of memory assignments for the \textbf{Star} experiment with $3$ groups.
    The values are normalized by the value of the best memoryless strategy found, optimum is zero.
    The optimal strategy requires more than a degree amount of memory.
    Automatic assignment achieves optimum consistently.}
    \label{fig-star}
\end{figure}

\paragraph{Stars}
We introduce a new benchmark on star-shaped graphs, where the targets and the attack lengths are chosen so that the optimal strategy requires a nontrivial amount of memory.
The graph is parametrized by a number of groups $k$ ranging from 1 to 5.
A graph with $k$ groups has $k + 2$ vertices.
In Figure~\ref{fig-star}, we report the statistics for $k=3$.

\begin{figure}[tb]
    \centering
    \includegraphics[width=\columnwidth]{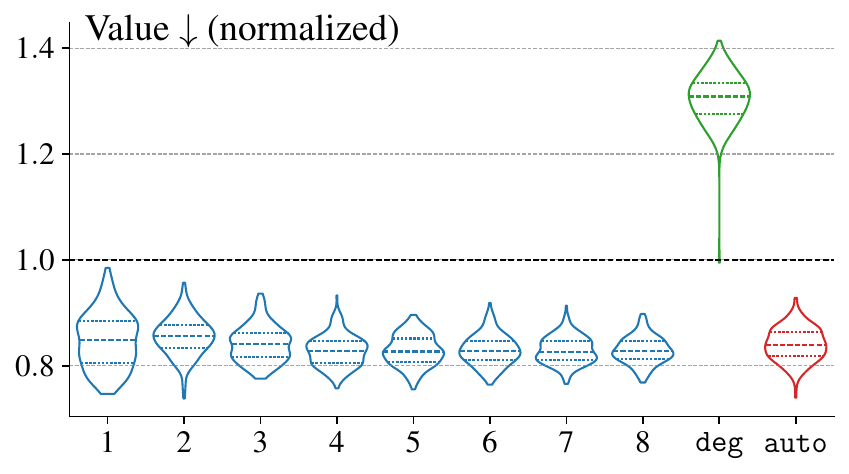}
    \caption{Comparison of memory assignments for the \textbf{Terrains} experiment with size $n=33$.
    The values are normalized by a baseline (the best deterministic strategy).
    Larger memory does not help here.
    Degree assignment causes the state space to be too large, and well-performing strategies are not found.}
    \label{fig-terrains}
\end{figure}

\paragraph{Terrains}
Another new benchmark consists of randomly generated connected planar graphs that model open terrains.
The nodes are randomly generated on a plane, triangulated, connected by the minimum spanning tree, and each edge from triangulation is added with a probability of 0.5.
We run the experiments with the odd number of nodes $n$ ranging from 3 to 41.
Figure~\ref{fig-terrains} shows the statistics for $n=33$.

\subsection{Discussion}
\label{sec:discussion}

\paragraph{Failure of uniform}
We observe that uniform memory assignments are rarely optimal.
Oftentimes, more memory is needed, and low values of $M$ achieve poor results.
In theory, if there is an optimal finite-memory strategy, then a high enough uniform assignment suffices.
The authors of \cite{KKMR:Regstar-UAI} state that the probability of finding the optimal strategy increases with $M$ in the Offices benchmark.
We verify this, but higher $M$ also significantly increases the state space and the time needed to find the optimal strategy (see Figure~\ref{fig:offices_time_plot}).

Therefore, within a fixed timeout, these two competing aspects (sizes of $M$ and state space) create a convex-shaped dependency on $M$, where the lowest values may or may not hit the optimum (cf.~Figures~\ref{fig-airport} and \ref{fig:office_floor_2}).
The behavior is hard to predict without intensive testing, making the uniform assignment impractical.

We can conclude that the distribution of memory states matters.
Degree assignment proved to be a simple heuristic that often yields superior results. 
Automatic assignment achieves better or comparable results than uniform in almost all cases.

\paragraph{Limitations of degree}
For Offices and Airport benchmarks, degree memory assignment produced the best results.
However, it scales poorly when it adds too much memory (Terrains, Building), see Figure~\ref{fig-terrains}.
On the opposite, it may provide an insufficient amount of memory, see Figure~\ref{fig-star}.
While it is a good heuristic for smaller graphs with a small maximal degree, it may achieve poor results on big or highly connected graphs or in cases when the optimal strategy requires more information than remembering the next vertex.  

\paragraph{Robustness of auto}
Our proposed automatic assignment achieves high robustness across all the experiments.
This heuristic enables the user to fully utilize the power of the synthesis tool.
On the one hand, we can push the tool to find complex strategies requiring memory (Offices, Airports, Stars).
On the other hand, we keep memory low if it is not needed, enabling the tool to converge to a high-quality strategy (Terrains).
Our method also scales well (Building).

A notable exception is the slight under-performance on the Airport benchmark, where for larger graphs, our automatic method achieves worse results than \texttt{deg}, see Figure~\ref{fig-airport}.
The reason is that this benchmark uses linear targets whose objective function is not bounded, and the convergence of positional strategies is slow.

We hypothesize that modifying the linear objective so that it is bounded would lead to better performance of \texttt{auto}, even for larger graphs.
The advantage of our approach is that our heuristic would be usable for this modification as well.

\section{Conclusion}

An automatic and robust memory assignment tool was a missing part of up-to-date, promising strategy synthesis techniques.
Our tool enables users and experts to fully utilize the power of the synthesis tools.

We proposed an automatic memory assignment method for patrolling problems and compared it against uniform and expert, degree-based assignments.
Our experiments across diverse benchmarks showed that our automatic method outperformed uniform memory in nearly all cases, offering robust and scalable performance.

Our approach provides a versatile and novel heuristic that is independent of the underlying optimization algorithms and objectives, making it applicable to a wider class of patrolling-type problems and new optimization techniques.

\section{Acknowledgments}
Vojtěch Řehák is supported by the Czech Science Foundation,  grant No.~26-23441S.
\bibliography{kur}

@book{Tambe:book,
	author = {Milind Tambe},
	title = {Security and Game Theory - Algorithms, Deployed Systems, Lessons Learned},
	publisher = {Cambridge University Press},
	year = {2012},
	url = {http://www.cambridge.org/de/academic/subjects/computer-science/communications-information-theory-and-security/security-and-game-theory-algorithms-deployed-systems-lessons-learned?format=AR},
	isbn = {978-1-10-709642-4},
	bibsource = {dblp computer science bibliography, https://dblp.org}
}

@article{ASYTBDMM:Protect-AIMagazine,
	author = {Bo An and
                  Eric Shieh and
                  Milind Tambe and
                  Rong Yang and
                  Craig Baldwin and
                  Joseph DiRenzo and
                  Ben Maule and
                  Garrett Meyer},
	title = {{PROTECT} - {A} Deployed Game Theoretic System for Strategic Security
                  Allocation for the United States Coast Guard},
	journal = {{AI} Mag.},
	volume = {33},
	number = {4},
	pages = {96--110},
	year = {2012},
	url = {https://doi.org/10.1609/aimag.v33i4.2401},
	doi = {10.1609/AIMAG.V33I4.2401},
	bibsource = {dblp computer science bibliography, https://dblp.org}
}

@inproceedings{ARSTMCC:multi-patrolling-survey,
  author       = {Alessandro Almeida and
                  Geber L. Ramalho and
                  Hugo Santana and
                  Patr{\'{\i}}cia Azevedo Tedesco and
                  Talita Menezes and
                  Vincent Corruble and
                  Yann Chevaleyre},
  tmpeditor    = {Ana L. C. Bazzan and
                  Sofiane Labidi},
  title        = {Recent Advances on Multi-agent Patrolling},
  booktitle    = {Proceedings of SBIA 2004},
  series       = {Lecture Notes in Computer Science},
  volume       = {3171},
  pages        = {474--483},
  publisher    = {Springer},
  year         = {2004},
  url          = {https://doi.org/10.1007/978-3-540-28645-5\_48},
  doi          = {10.1007/978-3-540-28645-5\_48},
  bibsource    = {dblp computer science bibliography, https://dblp.org}
}

@article{BGA:large-patrol-AI,
	author = {Nicola Basilico and
                  Nicola Gatti and
                  Francesco Amigoni},
	title = {Patrolling security games: Definition and algorithms for solving large
                  instances with single patroller and single intruder},
	journal = {Artif. Intell.},
	volume = {184-185},
	pages = {78--123},
	year = {2012},
	url = {https://doi.org/10.1016/j.artint.2012.03.003},
	doi = {10.1016/J.ARTINT.2012.03.003},
	bibsource = {dblp computer science bibliography, https://dblp.org}
}

@article{DJYZTKS:patrolling-uncertainty-JAIR,
	author = {Francesco Maria Delle Fave and
                  Albert Xin Jiang and
                  Zhengyu Yin and
                  Chao Zhang and
                  Milind Tambe and
                  Sarit Kraus and
                  John P. Sullivan},
	title = {Game-Theoretic Patrolling with Dynamic Execution Uncertainty and a
                  Case Study on a Real Transit System},
	journal = {J. Artif. Intell. Res.},
	volume = {50},
	pages = {321--367},
	year = {2014},
	url = {https://doi.org/10.1613/jair.4317},
	doi = {10.1613/JAIR.4317},
	bibsource = {dblp computer science bibliography, https://dblp.org}
}

@article{HZHH:multi-robot-patrol-survey,
	author = {Li Huang and
                  MengChu Zhou and
                  Kuangrong Hao and
                  Edwin S. H. Hou},
	title = {A survey of multi-robot regular and adversarial patrolling},
	journal = {{IEEE} {CAA} J. Autom. Sinica},
	volume = {6},
	number = {4},
	pages = {894--903},
	year = {2019},
	url = {https://doi.org/10.1109/JAS.2019.1911537},
	doi = {10.1109/JAS.2019.1911537},
	bibsource = {dblp computer science bibliography, https://dblp.org}
}

@inproceedings{PR:multi-patrolling-survey,
	author = {David Portugal and
                  Rui P. Rocha},
	title = {A Survey on Multi-robot Patrolling Algorithms},
	booktitle = {Proceedings of DoCEIS 2011},
	series = {{IFIP} Advances in Information and Communication Technology},
	volume = {349},
	pages = {139--146},
	publisher = {Springer},
	year = {2011},
	url = {https://doi.org/10.1007/978-3-642-19170-1\_15},
	doi = {10.1007/978-3-642-19170-1\_15},
	bibsource = {dblp computer science bibliography, https://dblp.org}
}

@inproceedings{AKK:multi-robot-perimeter-adversarial,
	author = {Noa Agmon and
                  Sarit Kraus and
                  Gal A. Kaminka},
	title = {Multi-robot perimeter patrol in adversarial settings},
	booktitle = {Proceedings of 2008 IEEE International Conference on Robotics and Automation},
	pages = {2339--2345},
	publisher = {{IEEE}},
	year = {2008},
	url = {https://doi.org/10.1109/ROBOT.2008.4543563},
	doi = {10.1109/ROBOT.2008.4543563},
	bibsource = {dblp computer science bibliography, https://dblp.org}
}

@inproceedings{AKKS:adversarial-uncertain,
	author = {Noa Agmon and
                  Sarit Kraus and
                  Gal A. Kaminka and
                  Vladimir Sadov},
	title = {Adversarial Uncertainty in Multi-Robot Patrol},
	booktitle = {Proceedings of IJCAI 2009},
	pages = {1811--1817},
	year = {2009},
	url = {http://ijcai.org/Proceedings/09/Papers/301.pdf},
	bibsource = {dblp computer science bibliography, https://dblp.org}
}

@inproceedings{BAVT:learn-preventive-healthcare,
	author = {Arpita Biswas and
                  Gaurav Aggarwal and
                  Pradeep Varakantham and
                  Milind Tambe},
	title = {Learn to Intervene: An Adaptive Learning Policy for Restless Bandits
                  in Application to Preventive Healthcare},
	booktitle = {Proceedings of IJCAI 2021},
	pages = {4039--4046},
	publisher = {ijcai.org},
	year = {2021},
	url = {https://doi.org/10.24963/ijcai.2021/556},
	doi = {10.24963/IJCAI.2021/556},
	bibsource = {dblp computer science bibliography, https://dblp.org}
}

@inproceedings{BGA:patrolling-arbitrary-topologies,
	author = {Nicola Basilico and
                  Nicola Gatti and
                  Francesco Amigoni},
	title = {Leader-follower strategies for robotic patrolling in environments
                  with arbitrary topologies},
	booktitle = {Proceedings of AAMAS 2009},
	pages = {57--64},
	publisher = {{IFAAMAS}},
	year = {2009},
	url = {https://dl.acm.org/citation.cfm?id=1558020},
	bibsource = {dblp computer science bibliography, https://dblp.org}
}

@inproceedings{BKKMNR:Patrolling-changing-UAI,
	author = {Tomáš Br{\'a}zdil and David Kla{\v{s}}ka and Antonín Ku{\v{c}}era and Vít Musil and
					Petr Novotn{\'y} and Vojtěch {\v{R}}eh{\'a}k},
	title = {On-the-fly adaptation of patrolling strategies in changing environments},
	booktitle = {Proceedings of UAI 2022},
	series = {Proceedings of Machine Learning Research},
	volume = {180},
	pages = {244--254},
	publisher = {{PMLR}},
	year = {2022},
	url = {https://proceedings.mlr.press/v180/brazdil22a.html},
	bibsource = {dblp computer science bibliography, https://dblp.org}
}

@inproceedings{FKDYT:PAWS,
	author = {Benjamin J. Ford and
                  Debarun Kar and
                  Francesco Maria Delle Fave and
                  Rong Yang and
                  Milind Tambe},
	title = {{PAWS:} adaptive game-theoretic patrolling for wildlife protection},
	booktitle = {Proceedings of AAMAS 2014},
	pages = {1641--1642},
	publisher = {{IFAAMAS/ACM}},
	year = {2014},
	url = {http://dl.acm.org/citation.cfm?id=2616103},
	bibsource = {dblp computer science bibliography, https://dblp.org}
}

@inproceedings{HO:UAV-problem-PSPACE,
	author = {Hsi{-}Ming Ho and
                  Jo{\"{e}}l Ouaknine},
	title = {The Cyclic-Routing {UAV} Problem is PSPACE-Complete},
	booktitle = {Proceedings of ETAPS 2015},
	series = {Lecture Notes in Computer Science},
	volume = {9034},
	pages = {328--342},
	publisher = {Springer},
	year = {2015},
	url = {https://doi.org/10.1007/978-3-662-46678-0\_21},
	doi = {10.1007/978-3-662-46678-0\_21},
	bibsource = {dblp computer science bibliography, https://dblp.org}
}

@inproceedings{KKLR:patrol-gradient,
	author = {David Kla{\v{s}}ka and Antonín Ku{\v{c}}era and Tomáš Lamser and Vojtěch {\v{R}}eh{\'{a}}k},
	title = {Automatic Synthesis of Efficient Regular Strategies in Adversarial
                  Patrolling Games},
	booktitle = {Proceedings of AAMAS 2018},
	pages = {659--666},
	publisher = {{IFAAMAS}},
	year = {2018},
	url = {http://dl.acm.org/citation.cfm?id=3237481},
	bibsource = {dblp computer science bibliography, https://dblp.org}
}

@inproceedings{KKMR:Regstar-UAI,
	author = {David Kla{\v{s}}ka and Antonín Ku{\v{c}}era and Vít Musil and Vojtěch {\v{R}}eh{\'{a}}k},
	title = {Regstar: efficient strategy synthesis for adversarial patrolling games},
	booktitle = {Proceedings of UAI 2021},
	series = {Proceedings of Machine Learning Research},
	volume = {161},
	pages = {471--481},
	publisher = {{AUAI} Press},
	year = {2021},
	url = {https://proceedings.mlr.press/v161/klaska21a.html},
	bibsource = {dblp computer science bibliography, https://dblp.org}
}

@inproceedings{KKR:patrol-drones,
	author = {David Kla{\v{s}}ka and Antonín Ku{\v{c}}era and Vojtěch {\v{R}}eh{\'{a}}k},
	title = {Adversarial Patrolling with Drones},
	booktitle = {Proceedings of AAMAS 2020},
	pages = {629--637},
	publisher = {{IFAAMAS}},
	year = {2020},
	url = {https://dl.acm.org/doi/10.5555/3398761.3398837},
	doi = {10.5555/3398761.3398837},
	bibsource = {dblp computer science bibliography, https://dblp.org}
}

@inproceedings{KL:patrol-regular,
	author = {Antonín Ku{\v{c}}era and Tomáš Lamser},
	title = {Regular Strategies and Strategy Improvement: Efficient Tools for Solving
                  Large Patrolling Problems},
	booktitle = {Proceedings of AAMAS 2016},
	pages = {1171--1179},
	publisher = {{ACM}},
	year = {2016},
	url = {http://dl.acm.org/citation.cfm?id=2937095},
	bibsource = {dblp computer science bibliography, https://dblp.org}
}

@inproceedings{KMZGA:moving_targets,
	author = {Jan Karwowski and
                  Jacek Mandziuk and
                  Adam Zychowski and
                  Filip Grajek and
                  Bo An},
	title = {A Memetic Approach for Sequential Security Games on a Plane with Moving
                  Targets},
	booktitle = {Proceedings of AAAI 2019},
	pages = {970--977},
	publisher = {{AAAI} Press},
	year = {2019},
	url = {https://doi.org/10.1609/aaai.v33i01.3301970},
	doi = {10.1609/AAAI.V33I01.3301970},
	bibsource = {dblp computer science bibliography, https://dblp.org}
}

@inproceedings{PJMOPTWPK:Deployed-ARMOR,
	author = {James Pita and
                  Manish Jain and
                  Janusz Marecki and
                  Fernando Ord{\'{o}}{\~{n}}ez and
                  Christopher Portway and
                  Milind Tambe and
                  Craig Western and
                  Praveen Paruchuri and
                  Sarit Kraus},
	title = {Deployed {ARMOR} protection: the application of a game theoretic model
                  for security at the Los Angeles International Airport},
	booktitle = {Proceedings of AAMAS 2008},
	pages = {125--132},
	publisher = {{IFAAMAS}},
	year = {2008},
	url = {https://dl.acm.org/citation.cfm?id=1402819},
	bibsource = {dblp computer science bibliography, https://dblp.org}
}

@inproceedings{SFAKT:Stackelberg-Security-Games,
	author = {Arunesh Sinha and
                  Fei Fang and
                  Bo An and
                  Christopher Kiekintveld and
                  Milind Tambe},
	title = {Stackelberg Security Games: Looking Beyond a Decade of Success},
	booktitle = {Proceedings of IJCAI 2018},
	pages = {5494--5501},
	publisher = {ijcai.org},
	year = {2018},
	url = {https://doi.org/10.24963/ijcai.2018/775},
	doi = {10.24963/IJCAI.2018/775},
	bibsource = {dblp computer science bibliography, https://dblp.org}
}

@inbook{TRKOT:IRIS,
    place={Cambridge},
    title={IRIS – A Tool for Strategic Security Allocation in Transportation Networks},
    booktitle={Security and Game Theory: Algorithms, Deployed Systems, Lessons Learned},
    publisher={Cambridge University Press},
    author={Tsai, Jason and Rathi, Shyamsunder and Kiekintveld, Christopher and Ordóñez, Fernando and Tambe, Milind and Tambe, Milind},
    year={2011}, pages={88–106}
}

@inproceedings{VAT:adversarial-patrolling,
	author = {Yevgeniy Vorobeychik and
                  Bo An and
                  Milind Tambe},
	title = {Adversarial patrolling games},
	booktitle = {Proceedings of AAMAS 2012},
	pages = {1307--1308},
	publisher = {{IFAAMAS}},
	year = {2012},
	url = {http://dl.acm.org/citation.cfm?id=2343977},
	bibsource = {dblp computer science bibliography, https://dblp.org}
}

@inproceedings{WSYWSJF:Patrolling-learning,
	author = {Yufei Wang and
                  Zheyuan Ryan Shi and
                  Lantao Yu and
                  Yi Wu and
                  Rohit Singh and
                  Lucas Joppa and
                  Fei Fang},
	title = {Deep Reinforcement Learning for Green Security Games with Real-Time
                  Information},
	booktitle = {Proceedings of AAAI 2019},
	pages = {1401--1408},
	publisher = {{AAAI} Press},
	year = {2019},
	url = {https://doi.org/10.1609/aaai.v33i01.33011401},
	doi = {10.1609/AAAI.V33I01.33011401},
	bibsource = {dblp computer science bibliography, https://dblp.org}
}

@inproceedings{Xu:Green-security,
	author = {Lily Xu},
	title = {Learning and Planning Under Uncertainty for Green Security},
	booktitle = {Proceedings of IJCAI 2021},
	pages = {4927--4928},
	publisher = {ijcai.org},
	year = {2021},
	url = {https://doi.org/10.24963/ijcai.2021/695},
	doi = {10.24963/IJCAI.2021/695},
	bibsource = {dblp computer science bibliography, https://dblp.org}
}

@inproceedings{YKKCT:Stackelberg-Nash-security,
	author = {Zhengyu Yin and
                  Dmytro Korzhyk and
                  Christopher Kiekintveld and
                  Vincent Conitzer and
                  Milind Tambe},
	title = {Stackelberg vs. Nash in security games: interchangeability, equivalence,
                  and uniqueness},
	booktitle = {Proceedings of AAMAS 2010},
	pages = {1139--1146},
	publisher = {{IFAAMAS}},
	year = {2010},
	url = {https://dl.acm.org/citation.cfm?id=1838360},
	bibsource = {dblp computer science bibliography, https://dblp.org}
}

@inproceedings{MunozdeCote2013,
	author = {Enrique Munoz de Cote and
                  Ruben Stranders and
                  Nicola Basilico and
                  Nicola Gatti and
                  Nick R. Jennings},
	title = {Introducing alarms in adversarial patrolling games: extended abstract},
	booktitle = {Proceedings of AAMAS 2013},
	pages = {1275--1276},
	publisher = {{IFAAMAS}},
	year = {2013},
	url = {http://dl.acm.org/citation.cfm?id=2485180},
	bibsource = {dblp computer science bibliography, https://dblp.org}
}

@article{SLESSLIN2019,
	author = {Efrat Sless and
                  Noa Agmon and
                  Sarit Kraus},
	title = {Multi-robot adversarial patrolling: Handling sequential attacks},
	journal = {Artif. Intell.},
	volume = {274},
	pages = {1--25},
	year = {2019},
	url = {https://doi.org/10.1016/j.artint.2019.02.004},
	doi = {10.1016/J.ARTINT.2019.02.004},
	bibsource = {dblp computer science bibliography, https://dblp.org}
}

@article{Basilico2022:recent-trends-in_rob_patrol,
	author = {Basilico, Nicola},
	title = {Recent Trends in Robotic Patrolling},
	journal = {Current Robotics Reports},
	year = {2022},
	volume = {3},
	number = {2},
	pages = {65--76},
	date = {2022-06-01},
	issn = {2662-4087},
	tempdoi = {10.1007/s43154-022-00078-5},
	url = {https://doi.org/10.1007/s43154-022-00078-5}
}

@inproceedings{KKMR:expected-intrusion-aamas,
	author = {David Klaška and
                  Antonín Kučera and
                  Vít Musil and
                  Vojtěch Řeh{\'{a}}k},
	title = {Minimizing Expected Intrusion Detection Time in Adversarial Patrolling},
	booktitle = {Proceedings of AAMAS 2022},
	pages = {1660--1662},
	publisher = {{IFAAMAS}},
	year = {2022},
	url = {https://www.ifaamas.org/Proceedings/aamas2022/pdfs/p1660.pdf},
	doi = {10.5555/3535850.3536068},
	bibsource = {dblp computer science bibliography, https://dblp.org}
}

@article{ACKL:patrol-in-uniform,
	author = {Steve Alpern and
                  Paul Chleboun and
                  Stamatios Katsikas and
                  Kyle Y. Lin},
	title = {Adversarial Patrolling in a Uniform},
	journal = {Oper. Res.},
	volume = {70},
	number = {1},
	pages = {129--140},
	year = {2022},
	url = {https://doi.org/10.1287/opre.2021.2152},
	doi = {10.1287/OPRE.2021.2152},
	bibsource = {dblp computer science bibliography, https://dblp.org}
}

@article{BDG:spatially-uncertain,
	author = {Nicola Basilico and
                  Giuseppe De Nittis and
                  Nicola Gatti},
	title = {Adversarial patrolling with spatially uncertain alarm signals},
	journal = {Artif. Intell.},
	volume = {246},
	pages = {220--257},
	year = {2017},
	url = {https://doi.org/10.1016/j.artint.2017.02.007},
	doi = {10.1016/J.ARTINT.2017.02.007},
	bibsource = {dblp computer science bibliography, https://dblp.org}
}

@inproceedings{CCGKK:fragmented-boundaries-memoryless,
	author = {Andrew Collins and
                  Jurek Czyzowicz and
                  Leszek Gasieniec and
                  Adrian Kosowski and
                  Evangelos Kranakis and
                  Danny Krizanc and
                  Russell Martin and
                  Oscar Morales{-}Ponce},
	title = {Optimal patrolling of fragmented boundaries},
	booktitle = {Proceedings of SPAA 2013},
	pages = {241--250},
	publisher = {{ACM}},
	year = {2013},
	url = {https://doi.org/10.1145/2486159.2486176},
	doi = {10.1145/2486159.2486176},
	bibsource = {dblp computer science bibliography, https://dblp.org}
}

@inproceedings{non-adversarial-patrolling1,
	author = {Peyman Afshani and
                  Mark de Berg and
                  Kevin Buchin and
                  Jie Gao and
                  Maarten L{\"{o}}ffler and
                  Amir Nayyeri and
                  Benjamin Raichel and
                  Rik Sarkar and
                  Haotian Wang and
                  Hao{-}Tsung Yang},
	title = {Approximation Algorithms for Multi-Robot Patrol-Scheduling with Min-Max
                  Latency},
	booktitle = {Proceedings of WAFR 2021},
	series = {Springer Proceedings in Advanced Robotics},
	volume = {17},
	pages = {107--123},
	publisher = {Springer},
	year = {2021},
	url = {https://doi.org/10.1007/978-3-030-66723-8\_7},
	doi = {10.1007/978-3-030-66723-8\_7},
	bibsource = {dblp computer science bibliography, https://dblp.org}
}

@article{non-adversarial-patrolling2,
	author = {Sai Krishna Kanth Hari and
                  Sivakumar Rathinam and
                  Swaroop Darbha and
                  Krishna Kalyanam and
                  Satyanarayana Gupta Manyam and
                  David W. Casbeer},
	title = {Optimal {UAV} Route Planning for Persistent Monitoring Missions},
	journal = {{IEEE} Trans. Robotics},
	volume = {37},
	number = {2},
	pages = {550--566},
	year = {2021},
	url = {https://doi.org/10.1109/TRO.2020.3032171},
	doi = {10.1109/TRO.2020.3032171},
	bibsource = {dblp computer science bibliography, https://dblp.org}
}

@article{non-adversarial-patrolling3,
	author = {Alessandro Farinelli and
                  Luca Iocchi and
                  Daniele Nardi},
	title = {Distributed on-line dynamic task assignment for multi-robot patrolling},
	journal = {Autonomous Robots},
	volume = {41},
	number = {6},
	pages = {1321--1345},
	year = {2017},
	url = {https://doi.org/10.1007/s10514-016-9579-8},
	tempdoi = {10.1007/S10514-016-9579-8},
	bibsource = {dblp computer science bibliography, https://dblp.org}
}

\clearpage

\appendix
\section{Appendix}

\section{Proof of the Sufficient Condition}

Generally, it can be seen that a sufficient condition on the first requirement of our framework is that the damage to targets depends only on the current state and the future.
That is, for every observation $o = (c_1, \ldots, c_n, c_{n+1})$ and target $\tau$ it holds
\begin{eqnarray}\label{eq-attack_no_history}
    \AttackVal(o, \tau \mid \sigma) = \AttackVal\bigl((c_n, c_{n+1}), \tau \mid \sigma\bigr)\ .
\end{eqnarray}
\begin{proof}
For the purposes of the proof, let us denote by $M$ the expression $\min_{B \in BSSC(\sigma)} \Val(\sigma \mid B)$.
We prove that for every strategy $\sigma$ on a set of states $C$ it holds
\begin{eqnarray*}
    \min_{c \in C}\,
          \sup_{\pi} \sum_{\substack{o \in \Omega \\ \pi(o) \in T}}\prob^{\sigma, c}\lbrack o \rbrack \cdot \AttackVal(o, \pi(o) \mid \sigma) \ = M.
\end{eqnarray*}
Also recall that the left-hand side is $\Val(\sigma) = \min_{c} \Val(\sigma, c) = \sup_{\pi} \Val(\sigma, c, \pi)$.

First, we will show $\Val(\sigma) \leq M$.
It is sufficient to show that there is a state $c \in C$, such that $\Val(\sigma, c) \leq M$.
Let us choose any $B \in BSCC(\sigma)$ such that $\Val(\sigma \mid B) = M$ and $c$ as any state in $B$.

Now, we aim to show that $\Val(\sigma, c, \pi) \leq \Val(\sigma \mid B)$ for every Attacker's strategy $\pi$.
Let $\pi$ be any strategy of the Attacker.
Using (\ref{eq-attack_no_history}) we can write that $ \Val(\sigma, c, \pi)$ equals
\begin{eqnarray*}
 \sum_{\substack{o \in \Omega \\ o = (c_1, \ldots, c_n, c_{n+1} ) \\ \pi(o) \in T}}\prob^{\sigma, c}\lbrack o \rbrack \cdot \AttackVal((c_n, c_{n+1}), \pi(o) \mid \sigma) \
\end{eqnarray*}
Since $c$ lies in a bottom strongly connected component $B$, every observation $o = (c_1, \ldots, c_n, c_{n+1})$ with positive $\prob^{\sigma, c}\lbrack o \rbrack$ must have that each transition $e = (c_i, c_{i+1})$ is in $B \times B$ and $\sigma(e) > 0$.
Namely $(c_n, c_{n+1}) \in B \times B$ and $\sigma(c_n, c_{n+1})> 0$.
From that it follows $ \AttackVal((c_n, c_{n+1}), \pi(o) \mid \sigma) \leq \Val(\sigma \mid B)$ for all observations with positive probability.
Using this, we can write
\begin{eqnarray*}
    \Val(\sigma, c, \pi) \leq  \sum_{\substack{o \in \Omega \\ \pi(o) \in T}}\prob^{\sigma, c}\lbrack o \rbrack \cdot \Val(\sigma \mid B) \\= \Val(\sigma \mid B) \sum_{\substack{o \in \Omega \\ \pi(o) \in T}}\prob^{\sigma, c}\lbrack o \rbrack \ .
\end{eqnarray*}
The desired inequality follows from the fact that the sum of probabilities of the observations where the Attacker attacks is at most $1$. Formally, it is true that
\begin{eqnarray*}
    \sum_{\substack{o \in \Omega \\ \pi(o) \in T}}\prob^{\sigma, c}\lbrack o \rbrack \leq 1 ,
\end{eqnarray*}
which follows from the fact that the Attacker is allowed to attack only once in a run.

For the second inequality $\Val(\sigma) \geq M$, it is sufficient to show that there is a strategy of the Attacker $\pi$ such that $\Val(\sigma, c, \pi) \geq M$  for every $c \in C$.

We now show how to construct such a strategy.
For every bottom strongly connected component $B$, choose any $\tau_{B} \in T$ and $e_B \in B \times B$ such that $\Val(\sigma \mid B) =  \AttackVal(e_{B}, \tau_{B} \mid \sigma)$ and $\sigma(e_{B}) > 0$.
Now, fix an observation $(c_1, \ldots, c_n, c_{n+1})$.
Let $i$ be the smallest index such that $(c_{i}, c_{i+1})$ is equal to some $e_B$. 
If no index has this property or $i < n$, then the Attacker waits.
Otherwise $(c_n, c_{n+1})$ is equal to a unique $e_B$ and the Attacker attacks $\tau_B$.

Let $c \in C$ be a state where the Defender starts patrolling.
The proof builds on two basic properties of discrete-time Markov chains.
No matter the starting state, the Defender will eventually visit some bottom strongly connected component $B$.
Here, the Defender will eventually visit $e_{B}$ with probability $1$ and the Attacker attacks $\tau_{B}$.
Since $M$ is chosen as the minimum over $\Val(\sigma \mid B)$, no matter which bottom strongly connected the Defender visits, the Attacker will achieve value at least $M$, which concludes the proof.
\end{proof}

\section{Strategy Expansion}

In \cite{KKMR:Regstar-UAI,BKKMNR:Patrolling-changing-UAI,KKMR:expected-intrusion-aamas}, the parameters of the strategy are randomly initialized at the start of the optimization.
However, it is also possible to warm-start from parameters provided by the user.
This may be useful because we add memory only to some locations.
Hence, it is possible that in the new strategy, some of the locations where we did not add memory could have the same optimal probabilities.
By warm-starting, we would preserve these probabilities, saving time in optimization.

For that reason, we describe how to create a new strategy $\sigma_2$ from $\sigma_1$ given the profile function resulting from \adjustmem{}.
For notation purposes, let $\mathrm{copies}(c) = |\mathrm{profiles(c)}|$.
The idea is to expand the strategy $\sigma_1$ such that every $c$ has now exactly $\mathrm{copies}(c)$ copies of $c$ 
that preserves the value of the strategy.

\begin{figure}[tb]\centering
\includegraphics[width=\columnwidth]{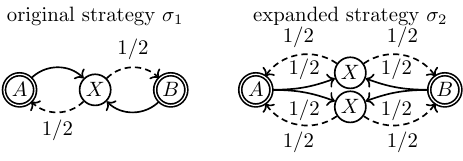}
\caption{Demonstration of strategy expansion. The original strategy is $\sigma_1$. The location $X$ gets two copies and is split, resulting in strategy $\sigma_2$.
The outgoing and incoming probabilities of both $X$s are identical. For both $X$s, the outgoing probabilities are preserved, but the incoming probabilities are halved between the two copies.}
\label{fig-expansion_simple}
\end{figure}

If $\mathrm{copies}(c) = 1$ for all but one state $X$ which has $\mathrm{copies}(X) = 2$, the procedure is simple.
We create an identical copy $X'$ of the state $X$.
That is, we add a new state with the same outgoing probabilities as $X$.
Then, every incoming edge to $X$ gets halved between $X$ and $X'$.
In Figure~\ref{fig-expansion_simple}, we show how this works on the positional strategy from the description of our method.

\begin{figure}[tb]\centering
\includegraphics[width=\columnwidth]{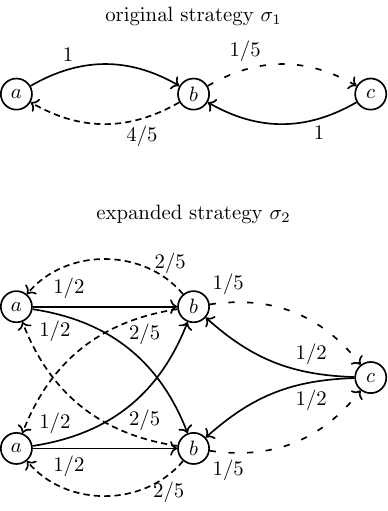}
\caption{Demonstration of strategy expansion. The original strategy $\sigma_1$ is memoryless.
Locations $a$ and $b$ both get two copies, resulting in expanded strategy $\sigma_2$.
The outgoing probabilities from $c$ and $a$ to $b$ are halved between the two $b$s (from probability $1$ to $1/2$).
The outgoing probability from $b$ to $a$ is halved between the two $a$s (from $4/5$ to $2/5$ to each copy). 
}
\label{fig-expansion}
\end{figure}

In general, let $C'$  denote the new set of states induced by the new memory assignment $\mem_2$.
For every state $c \in C$, let $c_1, \ldots, c_{\mathrm{copies}(c)}$ denote the corresponding new states in $C'$.
Each $c_i$ has the same distribution over $C'$ equal to
\begin{eqnarray}
    \sigma_2(c_i, x_j) = \frac{\sigma_1(c, x)}{\mathrm{copies}(x)}.
\end{eqnarray}
See Figure~\ref{fig-expansion} for a visual example in the general case.

In the experiment figures, we use \texttt{warm} to denote the runs with this variant of our method.

\section{Complexity Analysis}

Here, we describe the complexity analysis of our method, its bounded variation and strategy expansion.

\paragraph{Algorithm 1}
The time complexity of Algorithm 1 is in
\begin{equation*}
    \mathcal{O}(f + |\ellattacks| g + {|\ellattacks|}^2 {|C|}^2)    
\end{equation*}

where $f$ is such that the values of all attacks are calculated in $\mathcal{O}(f)$ and $g$ is such that for each attack, the gradients with respect to the parameters are evaluated in $\mathcal{O}(g)$.
Next, $|E|$ denotes the number of state edges such that its traversal probability is positive, and finally, $|\ellattacks|$ denotes the number of attacks taken into account.

\begin{proof}
In the algorithm, we keep a list of profiles for every state.
First, we need to determine the values of all attacks in $\mathcal{O}(f)$.
For every attack in the threshold, we calculate the gradients in $\mathcal{O}(g)$.
Then, for each state, we calculate the sign of the vector in $\mathcal{O}(|C|)$ and decide if it's already in the list.
This gives at most $|\ellattacks|$ iterations with equality of profiles being in $\mathcal{O}(|C|)$.
Determining the size of the profiles and summing over vertices is done in $\mathcal{O}(C)$.
\end{proof}
The time complexity of strategy expansion is evident.

\emph{In practice}, the evaluation for hard-constrained and blinded was faster than for linear. This is primarily because the first algorithm's core procedure is implemented in C++ with a dynamic program and uses a sparse representation of the strategy, unlike the second, which is in PyTorch and uses full matrices. In both cases, the actual bottleneck was in calculating the gradients.

\paragraph{Time Complexity} We claim that strategy expansion is in $\mathcal{O}(|C|^2)$, where $C$ is the \emph{new} number of states.
\begin{proof}
    For each new state $c$, evaluating the new outgoing probabilities can be easily done in $\mathcal{O}(|C|)$.
    This consists of copying the old probabilities divided by the number of new copies.
    This must be done for each state, hence in total $\mathcal{O}(|C|^2)$ operations.
\end{proof}

\paragraph{Bounded case} Let $M = \sum_{v}\mem_2(v)$. If $M \leq L$, then the algorithm is the same as the normal case.
Otherwise, we claim that in runs in
\begin{equation*}
    \mathcal{O}\left(|C||\ellattacks| + (M - |C|)\log(L - |C|)\right)
\end{equation*}
\begin{proof}
    Evaluating the value of all profiles of a single state can be implemented in $\mathcal{O}(|\ellattacks|)$.
    Hence, evaluating the value of all profiles of all states can be done in $\mathcal{O}(|C||\ellattacks|)$.
    The size of $\mathcal{P}$ is exactly $M - |C|$.
    Using heaps, extracting $L - |C|$ highest items of $\mathcal{P}$ takes $(M - |C|)\log(L - |C|)$.
\end{proof}

\section{Experiments}

Here, we offer the full report and description of our experiments.
In the code appended, we further describe how to recreate our experiments.
Moreover, we provide CSV files with the results of each experiment and the code used to generate the plots.

\subsection{Setup}

We ran our experiments on a server built on the Asus RS720A-E11-RS24U platform in the following configuration:
\begin{itemize}
    \item Two 64-core AMD EPYC 7713 2.0 GHz processors (128 physical cores and 256 threads in total);
    \item 2 TiB DDR4 RAM 3200 MHz;
    \item 10 Gbps Ethernet connection;
    \item 2 SATA SSDs with 960 GB capacity in RAID 1;
    \item 2 6 TB NVMe drives in RAID 1;
    \item Red Hat Enterprise Linux operating system.
\end{itemize}

Every experiment was run with lowered priority by command nice -n 10 and with a limit of $500$GB on usable RAM by system-md scope.

The optimization process can be thought of as a finite sequence of steps, which we call a \emph{run}.
In each step, we perform forward and backward passes, updating the parameters.
We executed $200$ different runs for every patrolling graph and memory (fixed memory assignment or our method).
The parameters were initialized randomly, and each run had a timeout of $180$ seconds.
In our method, we had bounded the number of states by $300$.

To explain the setup, we divide the run into \emph{epochs}.
An epoch is the maximal subsequence where the memory assignment stays constant.
For all baseline memory assignments, the whole run was a single epoch.
There were usually multiple epochs in the case of runs with our method.

\subsubsection{Terminating the optimization process}
If a single epoch took more than $2000$ steps, the whole run was terminated.
The run was also terminated if the value of the best strategy found was below $10^{-9}$ or if the run hit a plateau.

A \emph{plateau} occurs if, for $n$ steps, the value has not improved by the relative threshold of $\varepsilon$.
If the value of the best strategy found is $v$ and after $n$ steps, the best value is above $(1 - \varepsilon)v$, the run ends. We use values $n = 100$ and $\varepsilon = 10^{-5}$.
Plateau detection is taken into account only after $500$ steps of the current epoch.

\subsubsection{Changing the epochs}

Now, we describe how our method decides when to end an epoch and start a new one.
Each epoch was given at least $200$ steps.
After that, we use plateau detection but for $n = 20$ (threshold is the same).
If a plateau occurs during the epoch and the current epoch has improved the value by at least five percent, we change the memory according to the method, thus starting a new epoch.

Since both tools use Adam optimizer and noise, we naturally include that.
Starting a new epoch includes resetting the optimizer with the new parameters and resetting the noise.

\subsubsection{Hyperparameters of the tools}

The tools for solving the particular models also include various hyperparameters.
We use the parameters provided by the tools.
More details can be found in the project configuration files.
We consistently use the same hyperparameters across all experiments.

\subsection{Offices}

\begin{figure*}[tb]
    \centering
    \foreach \i in {1, 2, 3} {
        \subfigure[\i\ floors.]{
            \includegraphics[width=0.31\textwidth]{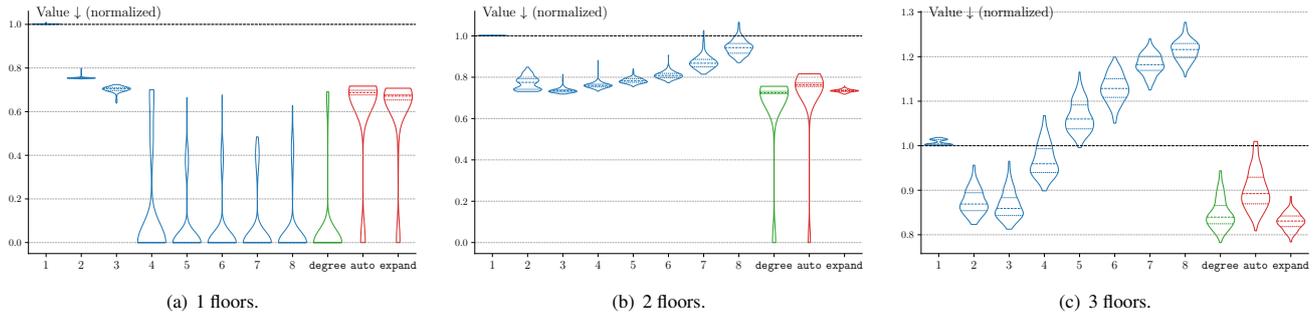}
            % \label{fig:offices_floor_\i}
        }
    }
    \caption{Statistics on the \textbf{Offices} benchmark with varying numbers of floors.}
    \label{fig:offices_floors_combined}
\end{figure*}

We evaluate our method on a benchmark introduced in \cite{KKMR:Regstar-UAI} that also appeared in \cite{BKKMNR:Patrolling-changing-UAI}.
The goal is to patrol offices in an office building.
The goal is to patrol offices (squares).
The corridors (circles) connect the offices on the same floor.
The floors are connected by staircases, which take $10$ units of time to traverse.
The traversal time between corridors is $2$, and the time between offices and corridors is $5$, as it involves opening the door and searching the office.

Here, all the targets (offices) are hard-constrained, with an attack length precisely equal to the shortest cycle visiting the entire building.
The offices have equal value.
The building is parametrized by the number of floors, and each floor has $4$ corridors that join the offices.
For short, we use $n$-floor to denote a building with $n$ floors.
Uniform memory assignment is used where every location is given $m$ memory values and $m$ ranges from $1$ to $8$.
We ran the experiments for buildings with one, two, and three floors.
In the original paper of \cite{KKMR:Regstar-UAI}, only a $1$-floor building is evaluated.

\subsection{Building}

We continue with the second benchmark on the office building topology, in which the targets are blinded.
This setting comes from \cite{KKMR:Regstar-UAI}.
The probability of discovering an ongoing attack upon each visit is $0.9$.
For buildings with $n$ floors, the attack lengths of all targets are equal to $100n$.
This is less than what is needed for the deterministic cycle, and the ratio to this length gets smaller with increasing $n$.
We ran the experiments for a number of floors running from $1$ to $5$.

\begin{figure*}[bt]
    \centering
    \foreach \i in {1, 2, 3, 4, 5} {
        \subfigure[\i\ floors.]{
            \includegraphics[width=0.31\textwidth]{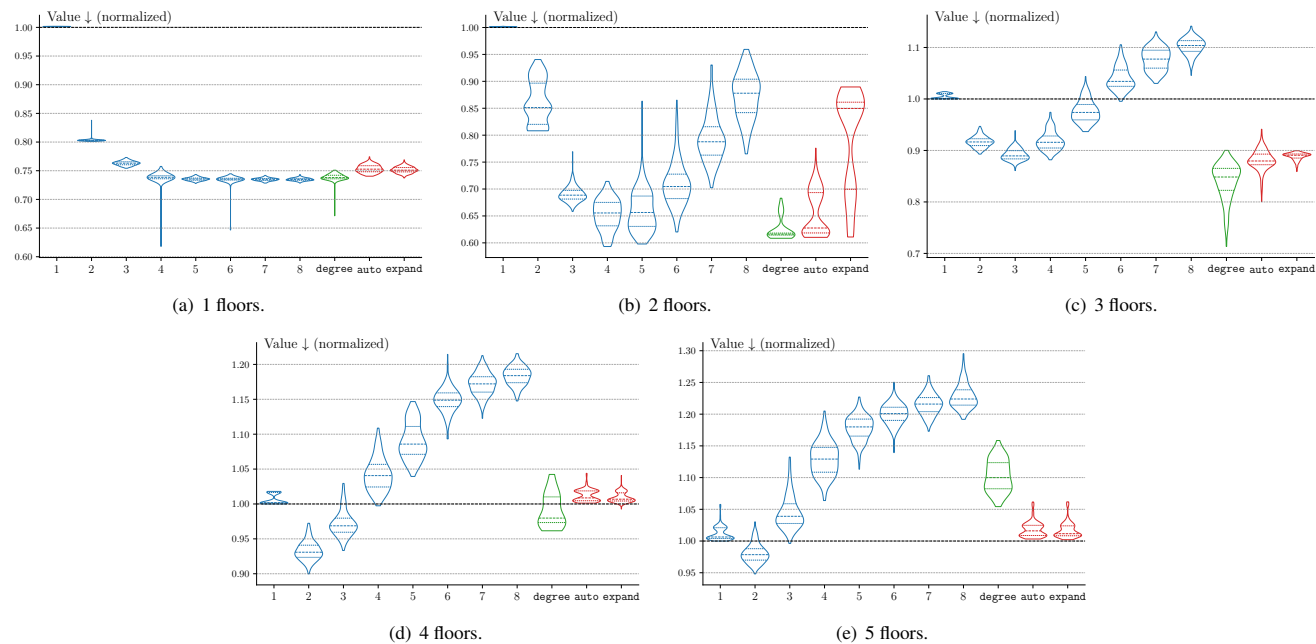}
        }
    }
    \caption{Statistics on the \textbf{Building} benchmark with varying number of floors.}
    \label{fig:building}
\end{figure*}

\subsection{Airports}

We evaluate our method on a model with linear targets.
This model has so far been studied only in \cite{KKMR:expected-intrusion-aamas}, where the authors use the following benchmark.

The goal is to patrol gates at an airport.
The airport consists of three terminals and $n$ halls.
Each hall is connected to two gates, and the airport forms a tree.
As a baseline, the authors use a strategy that cycles around the graph, visiting each target once on the cycle.
The value of such a strategy is then the length of this cycle ($3n + 1$ for the airport with $n$ halls) multiplied by the maximal value of gates.
The authors use the memory of the baseline strategy to run the optimization tool.
We denote this memory assignment as \texttt{deg}, as for each location, the memory is its degree in the graph.
To demonstrate the need for memory, we also report results with memoryless assignment $m = 1$.

In the setting used by the authors, all gates have the same unit value.
In that case, it is experimentally verified that the optimal strategy is actually the deterministic baseline.
Following the authors, we generate $10$ different airports, where the number of halls is a sequence 
\begin{eqnarray*}
        3,4,5,7,9,12,15,19,25,30 .
\end{eqnarray*}

We evaluated the methods in two different target settings.
One with the uniform values of gates and the second where each value is randomly and independently drawn from the interval $[1, 10]$.

\subsection{Stars}

Here, we introduce a benchmark \textbf{Stars} where the goal is to patrol the leaves of a star.
Let us denote by $S_k$ a graph in the topology of a star with $k$ leaves and internal location $M$.
The benchmark for a graph with group size $k$ is on a graph $S_{k + 1}$, where $d(v_1) = 4$ and $d(v_i) = 4k$ for $2 \leq i \leq k + 1$.
The defender needs to perform a cycle of length $4k$ of the form
\begin{eqnarray*}
    v_1 \to M \to v_2 \to M \to v_1 \to \cdots \to \\ v_1 \to M \to  v_k \to M \to v_1 \ .
\end{eqnarray*}
This requires $\mem(v_1) \geq k$ and $\mem(M) \geq 2k$.
The other targets need just one memory value.
We ran the experiments for $k$ ranging from $1$ to $5$.

The case for $k = 2$ is drawn in Figure~\ref{fig-deterministic_star}.

\begin{figure}[tb]
\centering
\includegraphics[width=\columnwidth]{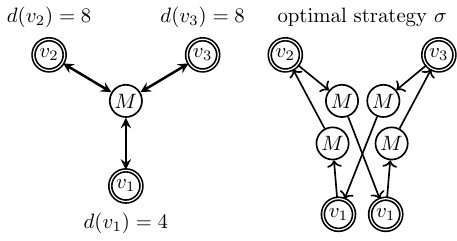}
\caption{Graph for \textbf{Stars} benchmark and $k=2$ (left). The strategy $\sigma$ achieves perfect protection since $v_1$ is always visited every $4$ steps, and both $v_2$ and $v_3$ are visited once on the cycle of length 8.
Strategy $\sigma_2$ needs $4$ states in $M$ and $2$ states in $v_1$.}
\label{fig-deterministic_star}
\end{figure}

\begin{figure*}[tb]
    \centering
    \foreach \i in {1, 2, 3, 4, 5} { 
        \subfigure[\i\ groups.]{
            \includegraphics[width=0.31\textwidth]{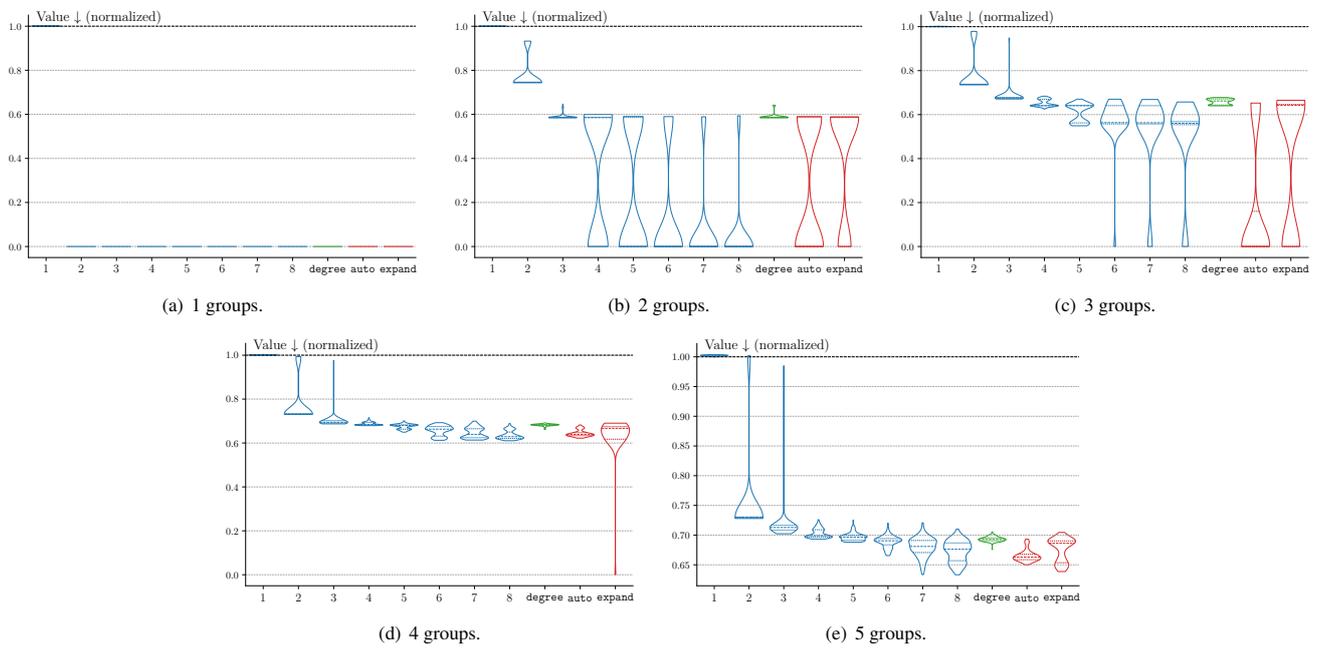}
            \label{fig:stars_groups_\i}
        }
    }
    \caption{Comparison of the methods on the \textbf{Stars} benchmark with varying numbers of groups.}
    \label{fig:stars_groups_combined}
\end{figure*}

\subsection{Terrains}

One use of patrolling games is in guarding large terrains (see Related Work).
Large terrains can be modeled as connected planar graphs, where each vertex represents a physical position in the terrain, and the length of edges is the time it takes to move between any two positions.

We compared the methods on a dataset of randomly generated connected planar graphs.
The construction for a graph on $n$ targets is as follows:
\begin{enumerate}
    \item Generate $n$ points uniformly in a unit 2D-box.
    \item Find the Delaunay triangulation $D$.
    \item Find a minimum spanning tree $T$ of $D$ where the length of edges is the Euclidean distance between the endpoints.
    \item Add each edge from $D$ to $T$ with probability $1/2$.
    \item Assign value to each target independently and uniformly from $[1, 10]$.
\end{enumerate}

We generated $20$ different graphs for $n$ ranging from $3$ to $41$ by steps of $2$.

As with the previous experiments on airports, we choose the strategy that cycles around the graph in the shortest possible time as a baseline.
Here, we use an approximate solution to the Traveling salesman problem.
For technical details on graph generation and the baseline, see the appended code.

\begin{figure*}[tbp]
    \centering
    \foreach \i in {3, 5, 7, 9, 11, 13, 15, 17, 19, 21, 23, 25} {
        \subfigure[Size \i.]{
            \includegraphics[width=0.31\textwidth]{figures/violin_plots/terrains_n_\i_violin.pdf}
            \label{fig:terrains_\i}
        }
    }
    \caption{Comparison of the method on the \textbf{Terrains} benchmark for varying sizes.}
    \label{fig:terrains_combined}
\end{figure*}

\begin{figure*}[tb]
    \centering
    \foreach \i in {27, 29, 31, 33, 35, 37, 39, 41} {
        \subfigure[Size \i.]{
            \includegraphics[width=0.31\textwidth]{figures/violin_plots/terrains_n_\i_violin.pdf}
            \label{fig:terrains_\i}
        }
    }
    \caption{Comparison of the method on the \textbf{Terrains} benchmark for varying sizes (continuation).}
    \label{fig:terrains_combined_large}
\end{figure*}

\subsection{Effect of Strategy Expansion}

From our experiments, we conclude that warm-starting the optimization process with strategy expansion is not significantly better.
A notable exception is \textbf{Stars} for $4$ groups.
Here, \texttt{expand} leads to the optimal value of $0$ as the only memory assignment.
The reason for this is due to an interesting phenomenon where the value stagnates and suddenly drifts away to the global minimum.
We observed this with constant memory assignment and on different graphs as well.
In the runs with strategy expansion, we give this phenomenon more steps to appear.
There are $20$ steps of patience before the plateau; then, we expand the strategy, and a new epoch begins.
Then, even if the value improves below the threshold and another begins, we are still virtually around the same local minima, albeit in a bigger state space.
This gives us at least $500$ additional steps before terminating the run.
Since the graph is small, they are executed before the timeout.

However, in general, finding the extreme values is more probable without warm-starting.
In cases where the memory needed is low, strategy expansion can lead to better convergence; this can be seen in \textbf{Terrains}.

\end{document}